\documentclass[10pt,twocolumn,letterpaper]{article}

\usepackage[pagenumbers]{cvpr} 

\newcommand\lft{\mathopen{}\left}
\newcommand\rgt{\aftergroup\mathclose\aftergroup{\aftergroup}\right}

\newcommand{\topic}[1]{\vspace{2mm}\noindent\textbf{#1.}}

\definecolor{cvprblue}{rgb}{0.21,0.49,0.74}
\usepackage[pagebackref,breaklinks,colorlinks,allcolors=cvprblue]{hyperref}

\def\paperID{****} 
\def\confName{CVPR}
\def\confYear{2025}

\title{Generative Multiview Relighting for \\ 3D Reconstruction under Extreme Illumination Variation}

\author{Hadi Alzayer\textsuperscript{\rm 1,2}~~~  Philipp Henzler\textsuperscript{\rm 1}~~~ Jonathan T. Barron\textsuperscript{\rm 1} \\ Jia-Bin Huang\textsuperscript{\rm 2}~~~ Pratul P. Srinivasan\textsuperscript{\rm 1}~~~ Dor Verbin\textsuperscript{\rm 1} \\
\\
\textsuperscript{\rm 1}Google \quad \quad
\textsuperscript{\rm 2}University of Maryland, College Park\\\\
\textbf{\url{https://relight-to-reconstruct.github.io/}}\\
}

\begin{document}
\twocolumn[{
\renewcommand\twocolumn[1][]{#1}
\maketitle
\begin{center}
\centering

\includegraphics[width=\linewidth]{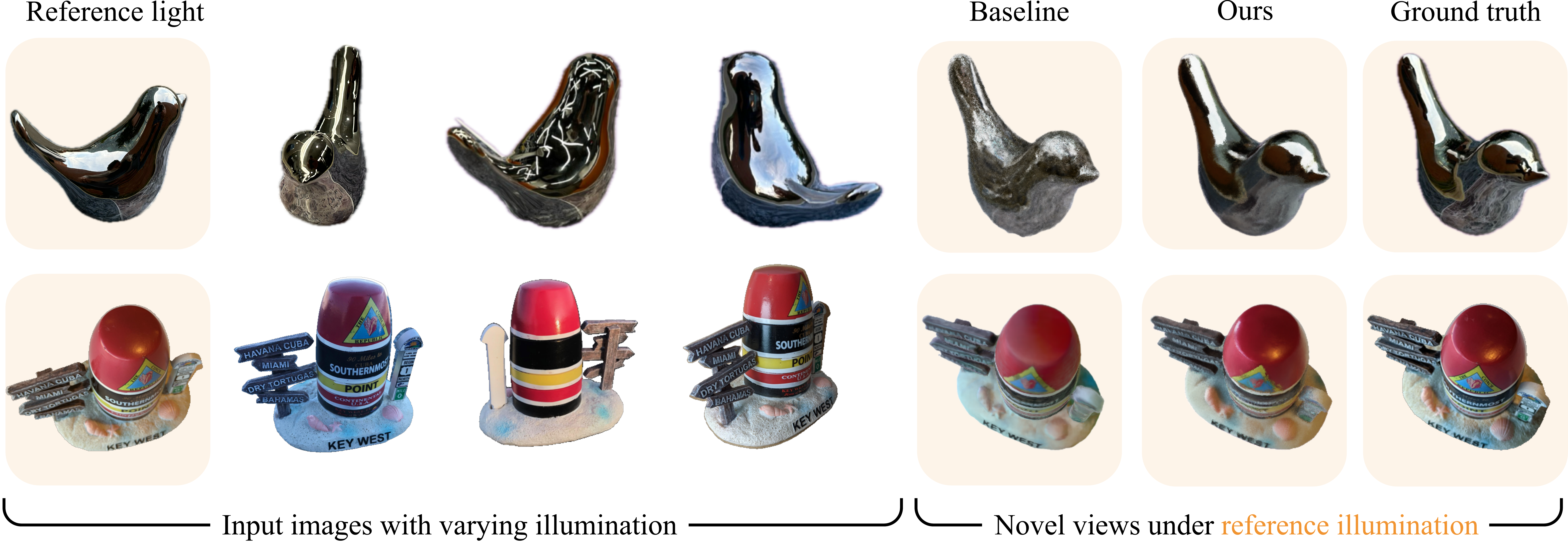}
\captionof{figure}{
\textbf{3D reconstruction under extreme illumination variation.} We propose a method for 3D reconstruction from a set of images captured under strongly varying illumination. Our method recovers high-fidelity appearance details including specular highlights that prior state-of-the-art approaches cannot recover (\emph{top baseline}: NeRF-Casting~\cite{verbin2024nerf} with appearance embeddings, \emph{bottom baseline}: NeROIC~\cite{neroic2022kuang}).
}

\label{fig:teaser}
\end{center}

}]

\begin{abstract}
Reconstructing the geometry and appearance of objects from photographs taken in different environments is difficult as the illumination and therefore the object appearance vary across captured images. This is particularly challenging for more specular objects whose appearance strongly depends on the viewing direction. Some prior approaches model appearance variation across images using a per-image embedding vector, while others use physically-based rendering to recover the materials and per-image illumination. Such approaches fail at faithfully recovering view-dependent appearance given significant variation in input illumination and tend to produce mostly diffuse results. We present an approach that reconstructs objects from images taken under different illuminations by first relighting the images under a single reference illumination with a multiview relighting diffusion model and then reconstructing the object's geometry and appearance with a radiance field architecture that is robust to the small remaining inconsistencies among the relit images. We validate our proposed approach on both synthetic and real datasets and demonstrate that it greatly outperforms existing techniques at reconstructing high-fidelity appearance from images taken under extreme illumination variation. Moreover, our approach is particularly effective at recovering view-dependent ``shiny'' appearance which cannot be reconstructed by prior methods. 

\end{abstract}    
\section{Introduction}
\label{sec:intro}

A common strategy for performing novel view synthesis is to recover a 3D representation from a collection of photographs of a scene, and then use it to render novel views from unseen viewpoints. 
Most view synthesis approaches assume that the input images are all taken under the same illumination conditions, \ie, the scene's appearance is assumed to be \emph{static}. 
However, this assumption is frequently violated during capture: moving clouds may cause the intensity of the sun to vary, artificial light sources may turn on or off, and even the shadows cast by the photographer may interfere with the scene.
This issue becomes even more severe when dealing with unstructured photograph collections such as those scraped from the Internet, where an object may be imaged in environments with extremely different illumination conditions, such as the case shown in Figure~\ref{fig:teaser}.

One strategy for addressing illumination variation is to use a view synthesis model capable of representing appearance changes across captured images. A simple and popular approach is to parameterize the view-dependent radiance of the scene as a function of a per-image ``latent code''~\cite{martinbrualla2020nerfw, bojanowski2018optimizing}. 
Such an approach is effective for modeling changes (\ie the ``base'' color and texture of the object). 
However, the additional expressivity comes at a cost because it tends to ``explain away'' \emph{all} view-dependent appearances (not just those due to variable illumination).
For non-diffuse (shiny) objects, we find that methods using per-image latent code often produce unrealistic, mostly-diffuse reconstructions.

Other physics-based approaches explicitly recover the spatially-varying material properties of the scene in addition to a per-image representation of illumination~\cite{boss2021nerd, neroic2022kuang}.
While inverse rendering provides a more physically meaningful image formation model than a view-dependent latent encoding, it suffers from the same ambiguities: appearance variation can be attributed to either a change in lighting or a change in viewing direction, and these models often mix up the two.

Recent works~\cite{jin2024neural_gaffer,zhao2024illuminerf} attempt to circumvent these ambiguities by training image diffusion models that relight an individual image to appear as if it were lit by a specified illumination condition. 
Though diffusion models can leverage strong priors on natural image appearance to assist in material/lighting disambiguation, single-image relighting is fundamentally highly ill-posed.
As such, independently applying these models to relight multiple images to a single reference illumination often yields samples corresponding to completely different and mutually-incompatible explanations of object materials.

In this work, we first use a \emph{multiview} image diffusion model to jointly relight all input images to match the lighting of a reference image selected from the set of input images. 
This produces relit images that are significantly more consistent than those from prior single-image relighting diffusion models. 
We then design a novel 3D reconstruction model based on NeRF~\cite{mildenhall2020nerf, verbin2024nerf} to fit a 3D representation to those relit images, which enables rendering from novel viewpoints. Our 3D reconstruction model is designed to be robust to inconsistencies that may still be present in the output of the relit samples. Since the diffusion model has to implicitly estimate 3D shape in order to relight the input images, any errors in its shape estimation would translate to specular highlights not appearing at the correct locations. To resolve this, we use a per-image vector which we call ``shading embedding''. This vector is used to encode per-image perturbations to the surface normal vectors used by the shading model of the NeRF, which allows the model to account for errors in the model's underlying estimated shape.

We validate the method on both synthetic (Objaverse~\cite{deitke2023objaverse}) and real (NAVI~\cite{jampani2023navi}) datasets.
We showcase the capability of reconstructing accurate geometry and accurate view-dependent appearance from images captured under extreme illumination variation. 
Extensive quantitative and qualitative evaluation shows that the proposed approach significantly outperforms state-of-the-art methods both qualitatively and quantitatively.
\section{Related Work}

\topic{3D Reconstruction and View Synthesis} 
Neural radiance fields (NeRF)~\cite{mildenhall2020nerf} is a successful approach for reconstructing a 3D scene under a fixed illumination. 
Many improvements to NeRF have been developed to improve the modeling of effects such as refractions and reflections~\cite{verbin2022refnerf,attal2022learning,bemana2022eikonal,liang2023envidr,verbin2024nerf,wu2024neural,ma2024specnerf}, but all of these approaches also rely on static lighting. 
NeRF-W~\cite{martinbrualla2020nerfw} achieved high-quality 3D reconstructions from unconstrained ``in the wild'' images taken from the internet, which tend to exhibit a variety of inconsistencies in lighting, appearance, and geometry. 
One aspect of this approach was the use of per-image appearance embeddings within the learned mapping from spatial coordinates and viewing direction to color (but not volumetric density). 
This parameterization forces optimization to recover a single consistent 3D model of geometry while allowing appearance to vary across input images.
This allows per-image appearance variation to be explained away, but also allows many view-dependent effects to be incorrectly attributed to per-image variation, often resulting in diffuse and erroneous reconstructions. 
Due to its simplicity (in that it does not contain a physically-based image formation model), per-image appearance embeddings are frequently used in settings where lighting only varies slightly~\cite{barron2022mipnerf360, barron2023zipnerf,tancik2022block}.

\topic{Inverse Rendering}
Decomposing an image into its constituent physical components has been a central problem in computer vision for nearly half a century~\cite{barrow1978recovering}. 
Most modern approaches heavily leverage the success of the computer graphics community, which has produced accurate and efficient models for the ``forward'' problem of rendering an image from an underlying 3D model~\cite{pharr2023physically}. 
Many modern inverse rendering techniques use radiance fields that, rather than mapping a 3D location and viewing direction to an outgoing color, map a 3D location to material properties and surface normals, which are then rendered according to some estimate of incident illumination~\cite{jin2023tensoir, boss2021nerd, boss2021neuralpil, engelhardt2023-shinobi, liu2023nero, mai2023neural,sun2023neural,nerv2021,attal2024flash,nrf2020}.
Because this problem is inherently ill-posed, these techniques often critically rely on analytical priors to regularize the estimated scene decomposition. 
Although these models fully decompose the scene into its components and thereby enable editing and relighting, their rigidly specified rendering models tend to constrain the solution space of the recovered model and result in lower rendering quality than models designed solely for view synthesis. 

Many prior approaches to inverse rendering depend critically on modeling illumination and material properties using low-frequency parameterizations such as spherical harmonics~\cite{neroic2022kuang}, spherical Gaussians~\cite{physg2021}, or MLPs that are strongly biased to produce smoothly-varying outputs~\cite{zhang2021nerfactor}. 
All of these parametrizations necessarily limit the model's ability to recover and render high-frequency (\ie, non-smooth) color variations caused by glossy surfaces. 
This limitation is avoided by NeRF-Casting~\cite{verbin2024nerf}, which models incident illumination by explicitly reflecting secondary rays into the volumetric representation, thereby enabling the reconstruction of extremely specular objects.
The 3D representation used by our model builds upon NeRF-Casting.

\topic{Lighting Estimation}
Because estimating every aspect of a 3D scene is a challenging task with a broad scope, many prior works have instead attempted to only reconstruct a model of illumination from input images~\cite{lalonde2012estimating,lighthouse2020,gardner2017,garon2019}. 
Though this problem is still fundamentally underconstrained, it can be made tractable by learning a model from datasets containing light probes~\cite{legendre2019deeplight} or by relying on accidental light probes in the scene~\cite{yu2023alp}. 
Modern generative approaches in this vein have shown that diffusion models are capable of developing an internal understanding of illumination~\cite{Phongthawee2023DiffusionLight}, which is further validated by our work. 

\topic{Generative Relighting}
Diffusion-based generative models have recently been used for relighting individual images~\cite{zeng2024dilightnet} and, closer to our approach, for 3D relighting~\cite{jin2024neural_gaffer,zhao2024illuminerf, irnia_relighting_2024}. 
This process circumvents the need to decompose the scene into physically-meaningful components and instead relies on a diffusion model to re-synthesize an input image as if it were illuminated by some reference illumination. Unlike our method, these approaches require the set of input images to have constant illumination, which is used in the first step to recover the 3D geometry of the object before relighting it. 
IllumiNeRF~\cite{zhao2024illuminerf} samples a diffusion model multiple times to generate a collection of plausible results, while Neural Gaffer~\cite{jin2024neural_gaffer} iteratively refines a set of relit images alongside a NeRF. 
Crucially, both models \emph{independently} relight each input image given a reference illumination, which can lead to inconsistent relit images. Additionally, both methods require a reference environment map as input, while we address a more general problem statement where we only have access to images.
Our model works by relighting a set of differently-lit input images \emph{simultaneously} to match the appearance of one of the input images, and it does not require any additional information, including constantly-lit input images or a reference environment map. We show that our approach not only yields significantly more consistent 3D reconstructions than prior work, but is also more practically applicable to a broader variety of inputs and capture settings.

\section{Method} \label{sec:method}



\begin{figure}
\begin{center}
\centering

\includegraphics[width=1.0\linewidth, trim=0 0 0 0, clip]{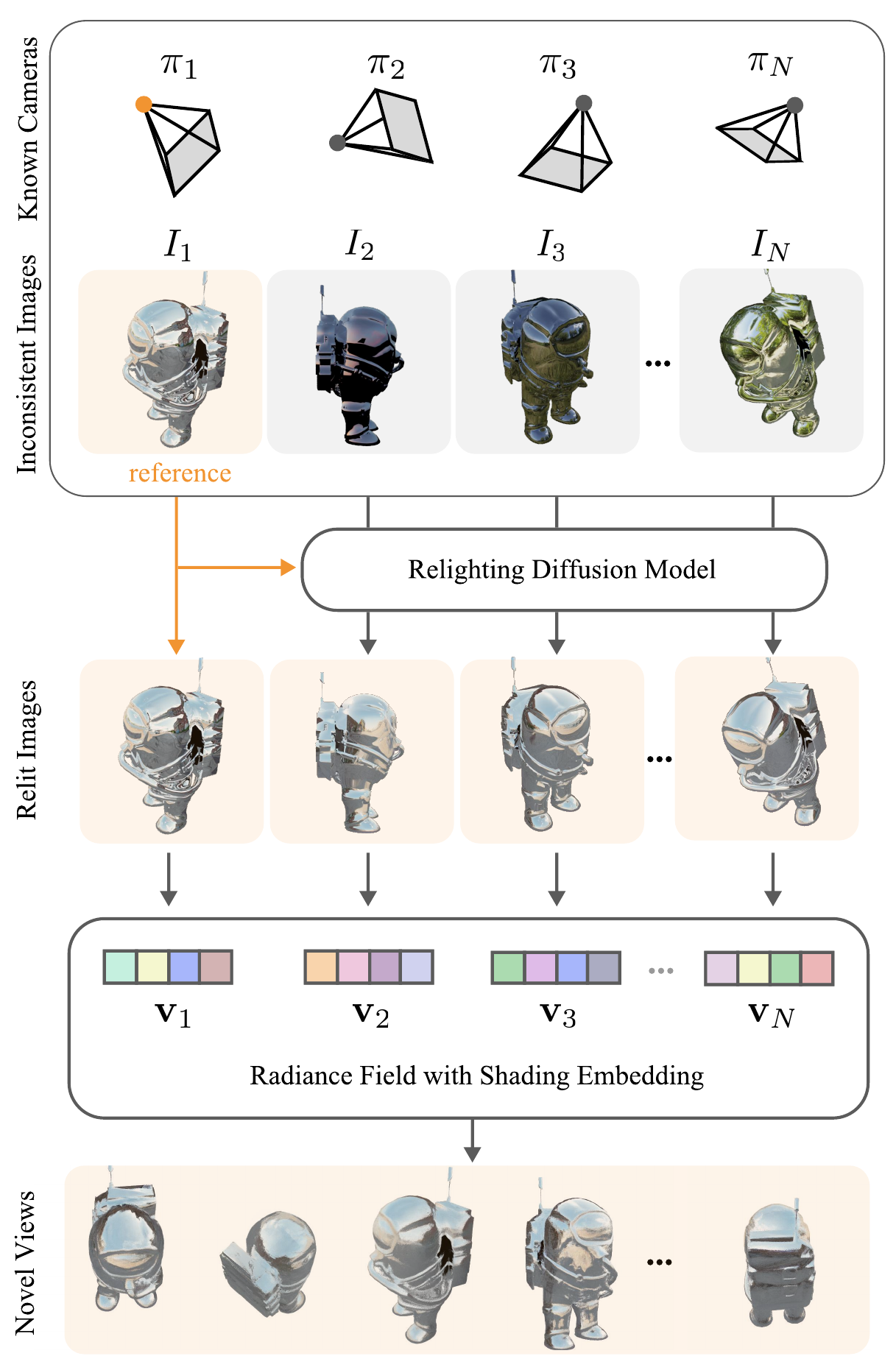}

\captionof{figure}{
\textbf{Method overview.} We first apply a relighting diffusion model that converts $N$ images $I_1$, ..., $I_N$ with known camera poses $\pi_1$, ..., $\pi_N$, captured under extremely different illuminations, to a set of images with the same poses, but rendered under the illumination of the reference image $I_1$ (highlighted in orange). We then optimize a neural radiance field to obtain a consistent 3D representation with a novel per-image shading embedding, which can be used to render new views of the scene from unobserved poses.
}
\label{fig:overview}
\end{center}
\end{figure}

We aim to recover a 3D model from an collection of (posed) images of an object, where each image is illuminated by arbitrarily-varying lights. We use a two-step solution: in Section~\ref{sec:relighting} we describe a relighting diffusion model that ``harmonizes'' the input images by making their illumination constant, and in Section~\ref{sec:nerf} we describe a NeRF-based approach to convert the relit images produced by that diffusion model into a consistent 3D model. Our full pipeline is illustrated in Figure~\ref{fig:overview}.

\subsection{Relighting model} \label{sec:relighting}

Our relighting model takes a set of $N$ input images $I_1, ..., I_{N}$ of an object from multiple known camera poses $\pi_1, ..., \pi_{N}$, each of which is assumed to have been taken under a different illumination (see Figure~\ref{fig:overview}, leftmost pane). The relighting model's aim is to generate a set of images of the same object from the same viewpoints, relit under the illumination from a designated reference image in the input set, which we arbitrarily select as the first image, $I_1$.

We build our relighting model using image diffusion, which can leverage strong natural image priors from large amounts of data. While recent works have used image diffusion models to relight individual input images based on a reference environment map~\cite{zeng2024dilightnet,zhao2024illuminerf,jin2024neural_gaffer}, we argue that it is essential to jointly relight the entire set of input images. As has been demonstrated by the success of photometric stereo~\cite{photometricstereo}, leveraging multiple observations of the same material under varying lights reduces the inherent ambiguity between material properties and illumination, and provides a strong cue for estimating shape. Using this same insight in the context of modern diffusion models leads to significantly more consistent relit images, as each sample from our model represents a unified interpretation of the object's material. 
Consequently, our method does not require ad-hoc solutions to address the issue of extremely-inconsistent relit images, such as the use of appearance embeddings~\cite{zhao2024illuminerf} or refinement of the relit images during reconstruction~\cite{jin2024neural_gaffer}.

\newcommand{\trimaa}{2.2in}
\newcommand{\trimab}{1.4in}
\newcommand{\trimac}{2.0in}
\newcommand{\trimad}{0.5in}

\newcommand{\trimba}{0.7in}
\newcommand{\trimbb}{0.8in}
\newcommand{\trimbc}{0.4in}
\newcommand{\trimbd}{2in}

\newcommand{\figthreewidth}{0.18\linewidth}

\begin{figure*}[t]
    \centering
    {
    \setlength\arrayrulewidth{1pt}
    \begin{tabular}{@{}c@{\,\,}c@{\,\,}|@{\,\,}c@{\,\,}@{\,}c@{\,\,}c@{}}
    \includegraphics[width=\figthreewidth, trim={\trimaa} {\trimab} {\trimac} {\trimad}, clip]{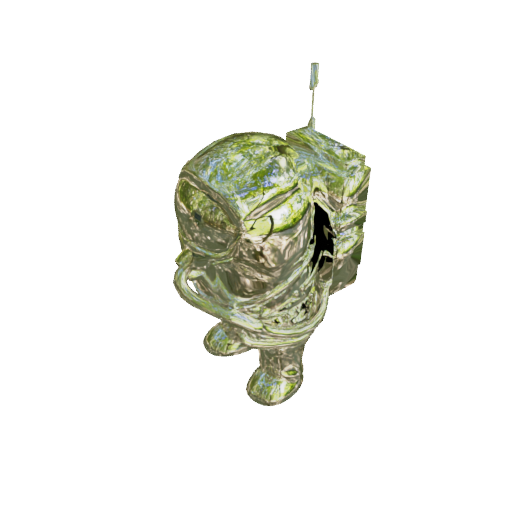} & 
    \includegraphics[width=\figthreewidth, trim={1.55in} {\trimab} {\trimac} {0in}, clip]{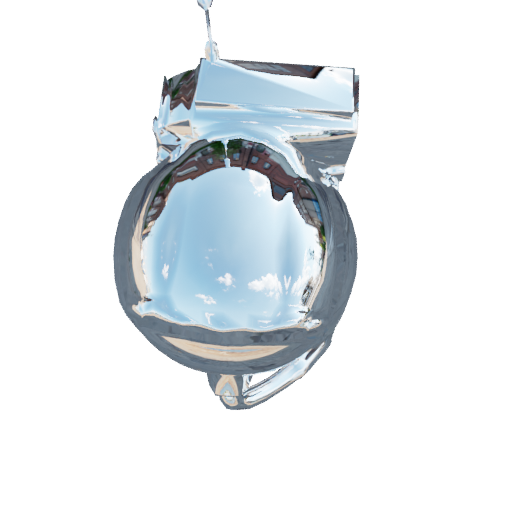} & 
    \includegraphics[width=\figthreewidth, trim={\trimaa} {\trimab} {\trimac} {\trimad}, clip]{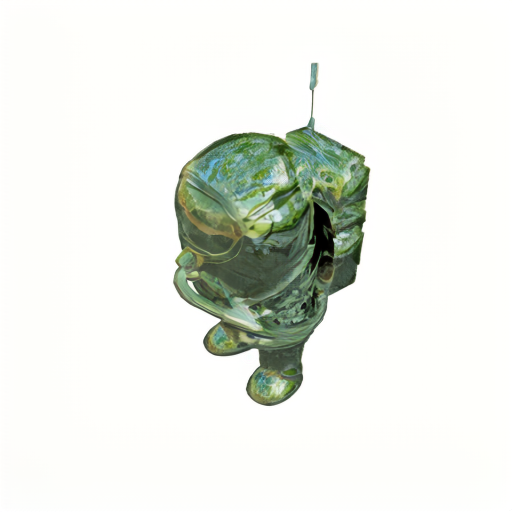} & 
    \includegraphics[width=\figthreewidth, trim={\trimaa} {\trimab} {\trimac} {\trimad}, clip]{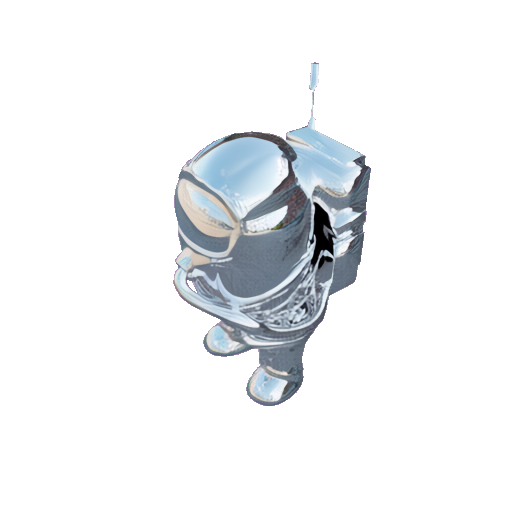} & 
    \includegraphics[width=\figthreewidth, trim={\trimaa} {\trimab} {\trimac} {\trimad}, clip]{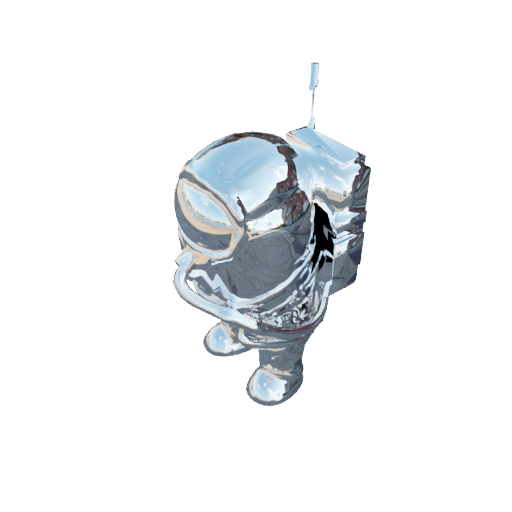} \\
    \includegraphics[width=\figthreewidth, trim={\trimba} {\trimbb} {\trimbc} {\trimbd}, clip]{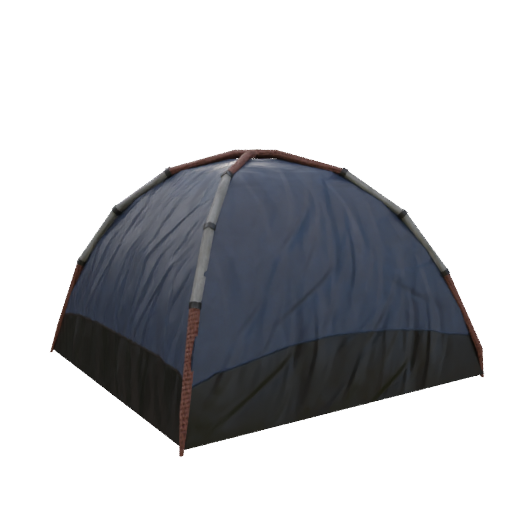} & 
    \includegraphics[width=\figthreewidth, trim={0.3in} {0.4in} {0in} {\trimbd}, clip]{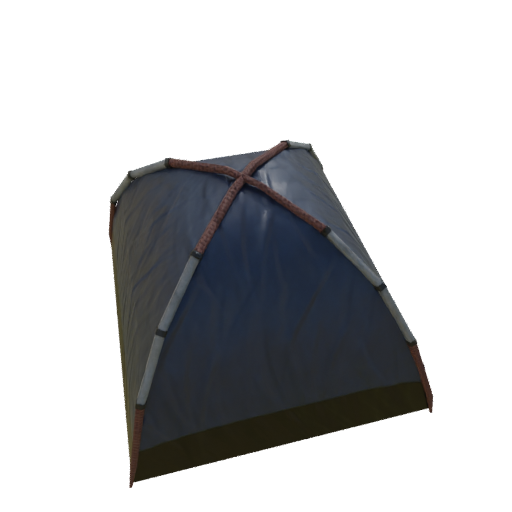} & 
    \includegraphics[width=\figthreewidth, trim={\trimba} {\trimbb} {\trimbc} {\trimbd}, clip]{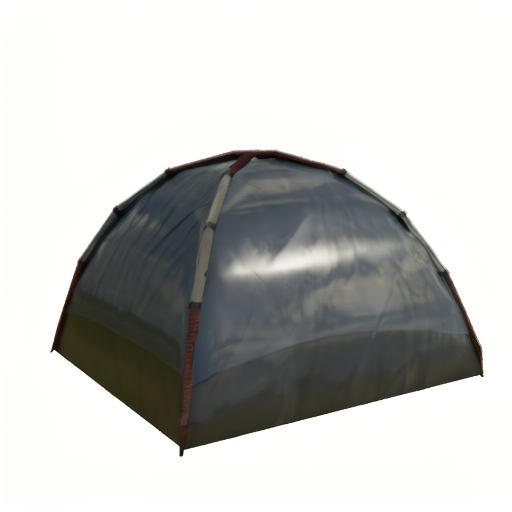} & 
    \includegraphics[width=\figthreewidth, trim={\trimba} {\trimbb} {\trimbc} {\trimbd}, clip]{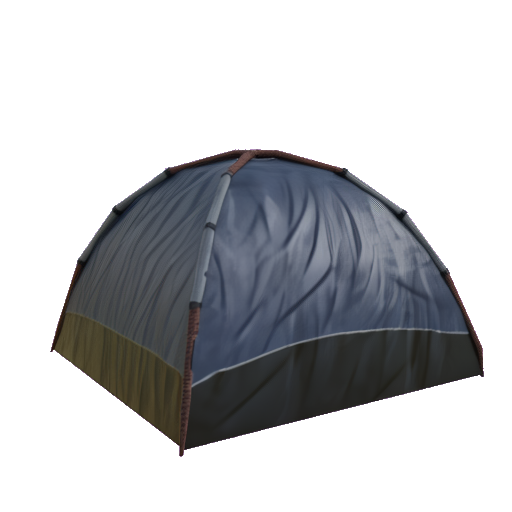} & 
    \includegraphics[width=\figthreewidth, trim={\trimba} {\trimbb} {\trimbc} {\trimbd}, clip]{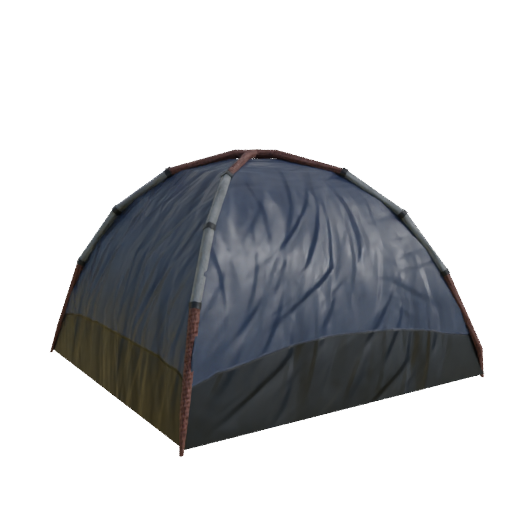} \\
    \small (a) Input Image to be Relit  & \small (b) Reference Image & \small (c)  IllumiNeRF~\cite{zhao2024illuminerf} & \small (d)  Our Output  & \small (e)  Ground Truth
    \end{tabular} 
    }
    \caption{
    \textbf{A comparison of our multiview relighting with prior work on single-image relighting.} Our method first relights a set of inconsistently-lit images (one of which is shown in (a)) to match the illumination of a selected reference image (b) in that set. Single-image relighting techniques such as IllumiNeRF~\cite{zhao2024illuminerf} (c) struggle to disambiguate geometry, lighting, and materials, leading to an inaccurate relighting. In contrast, our model jointly relights a set of inconsistently-lit frames, which reduces ambiguities and results in a significantly more accurate result (d) when compared with the ground truth (e). 
    }
    \label{fig:relighting_comparison}
\end{figure*}

Our model's architecture follows prior work on view synthesis that extends image generative models to the multiview or video case~\cite{ho2022video, blattmann2023align, gao2024cat3d} --- we will analogously extend a single-image relighting model to multiple views, each with a distinct light.

We construct our relighting diffusion model as a multiview latent model that denoises a set of $N$ latent codes $\mathbf{z}_1, ..., \mathbf{z}_{N}$, one for each input image.
Unlike generative models that synthesize images from only random noise, in our case the observed images $I_1, ..., I_{N}$ contain a significant amount of valuable information about geometry and illumination. As such, when denoising the $i$th latent code, $\mathbf{z}_i$, the denoising network should use the geometry encoded in image $I_i$ as well as the lighting of the reference image $I_1$. To accomplish this, we pass the ``clean'' latent corresponding to the $i$th image $I_i$ and an encoding of its pose $\pi_i$ into the network that denoises the $i$th latent $\mathbf{z}_i$. To jointly reason about the reference lighting, we use 3D self-attention blocks~\cite{gao2024cat3d} for the latent codes, and we use cross-attention in between them and the camera poses.  This structure allows the geometry and camera pose of the $i$th image as well as the lighting of the reference image to be directly observed by the network denoising the $i$th latent $\mathbf{z}_i$.

Because our input is an unordered set of images, we follow CAT3D's approach~\cite{gao2024cat3d} of replacing temporal embeddings with camera poses which we encode as raymaps~\cite{watson2022novel, sajjadi2022scene}.
To inform the network which image should be used as the reference, we concatenate the reference image with a ``reference map'' consisting of a single-channel image of ones, and we concatenate all other images with a single-channel image of zeros. See the supplement for a full description of this process. Results from this relighting model component are shown in Figure~\ref{fig:relighting_comparison}.

\subsection{3D Reconstruction} \label{sec:nerf}

Applying the relighting model from Section~\ref{sec:relighting} to the input images
yields a set of ``harmonized'' input images that appear to have been lit under a single consistent illumination. With these images we can recover a 3D model without solving the difficult problem of handling variable illumination within a 3D reconstruction pipeline.
We build our reconstruction model on NeRF-Casting~\cite{verbin2024nerf}, a view synthesis technique that achieves state-of-the-art results for reconstructing highly specular objects under constant illumination.

Although the relit images produced by our relighting diffusion model appear largely consistent when visually inspecting individual images, they still exhibit slight inconsistencies with each other.
As we will demonstrate, if the outputs of our diffusion model are used to naïvely train a NeRF that is not robust to this subtle variation (such as the common appearance-embedding based approach of NeRF-W~\cite{martinbrualla2020nerfw}), these inconsistencies may be absorbed into the view-dependent parameterization of the NeRF, which results in novel view synthesis results with highly unrealistic and distracting ``flickering'' temporal artifacts as the camera moves.
We must therefore be careful to parameterize appearance such that optimization is robust to this variation  --- resolving these inconsistencies is a primary goal of 3D reconstruction. 

We address this problem by observing that the subtle inter-image errors made by the diffusion model are usually due to the specular highlights being slightly ``warped'' relative to the ground truth. This suggests that the diffusion model has small errors in its implicit estimate of the surface normals of the object, which causes the specular reflections to be tilted in slightly incorrect directions. We therefore resolve this by using a per-image ``shading embedding'': an embedding vector $\mathbf{v}_i$ is used to warp the normal vectors of the corresponding $i$th training image, and those warped normal vectors are provided as input to the network that models the shading (and therefore, appearance) of the image. More concretely, the normals of the $i$th image at a 3D coordinate $\mathbf{x}$ are computed as:
\begin{equation} \label{eq:shadingnormals}
    \mathbf{n}_i(\mathbf{x}) = \operatorname{normalize}\lft(\operatorname{MLP}\lft(\mathbf{f}(\mathbf{x}), \mathbf{v}_i\rgt)\rgt)\,,
\end{equation}
where $\operatorname{MLP}(\cdot)$ is a multi-layer perceptron with $3$ layers each with $128$ hidden units, $\mathbf{f}(\mathbf{x})$ is a spatially-varying feature parameterizing the normals (identical to the one used by NeRF-Casting), and the $\operatorname{normalize}(\cdot)$ operator turns the output of the MLP to a unit-length vector.

The per-image surface normals defined in Equation~\ref{eq:shadingnormals} can slightly alter the directions at which secondary rays are reflected by the model, which allows optimization to account for errors in the output of the relighting diffusion model. We prevent the normals from changing significantly between different images by tying them to the normals corresponding to the underlying NeRF density field, as was done in NeRF-Casting~\cite{verbin2024nerf}. See NeRF-Casting and our supplement for more details.
As we will demonstrate later, this shading embedding yields more accurate reconstructions and renderings than the standard appearance embedding approach~\cite{martinbrualla2020nerfw} used in prior work for relighting~\cite{jin2024neural_gaffer,zhao2024illuminerf}.

\section{Architecture and Training} \label{sec:training}

\topic{Training Data}
The relighting model's training dataset plays a key role in its ability to generalize across different illumination conditions, materials, and geometries. Due to the ease of rendering a large set of synthetic data, we rely solely on synthetic data for training, and we experimentally verify that our method generalizes to real captured photographs.
We render our dataset using assets from a dataset of $\sim\!300$K high-quality objects, similar in appearance to Objaverse but with more diverse materials. We additionally augment this dataset by rendering another copy of it after replacing all materials with perfect mirrors, which we found improved reconstruction quality for both highly-specular objects and (perhaps surprisingly) for standard mostly-diffuse objects.
We render the objects using about $700$ environment maps taken from Poly Haven~\cite{zaal2021polyhaven}, which we augment by randomly rotating azimuthally.

\topic{Diffusion Model}
We initialize our model using an image generation latent diffusion model similar in architecture to Stable Diffusion 1.5~\cite{Rombach_2022_CVPR} that had been trained on a large dataset of images. 
The image input of our diffusion model is a $512 \times 512 \times 3$ resolution image that is encoded into $64 \times 64 \times 8$ latents. 
To enable classifier-free guidance (CFG)~\cite{ho2022classifier}, we randomly mask out the attention to the reference image to force the model to do ``unconditional relighting'' and relight the input views with an arbitrary lighting. For all our experiments, we use a CFG value of 3.
Because the speed and memory requirements of training the model are inversely proportional to the number of synthesized views, we train our model in stages: We first train the model to relight 8 frames for $200$K steps, then 16 frames for $100$K steps, then 32 frames for $50$K steps, and finally 64 frames for another $50$K steps. Note that our model parameters are independent of the number of input views, since the cross-attention operation can scale to an arbitrary number of images.
Our model was trained on 64 5th-generation TPUs with a total batch size of 64, which took approximately two weeks. At inference, we set the number of images that our diffusion model simultaneously relights to $N=64$

\topic{3D Reconstruction} The optimization of each radiance field uses a similar number of parameters to those used in NeRF-Casting~\cite{verbin2024nerf}. Optimizing a model from a set of harmonized images takes roughly $30$ minutes on 16 NVIDIA A100 GPUs, and rendering a single $512\times 512$ image takes about $0.5$ seconds on the same hardware. See supplement for details and a full description of all hyperparameters.
\section{Results}

We evaluate our method on two datasets: synthetic objects from Objaverse~\cite{deitke2023objaverse} and real captured objects from NAVI~\cite{jampani2023navi}. The synthetic dataset is made up of two components: $8$ standard textured assets whose materials range from mostly-diffuse to glossy, and $12$ shiny assets containing objects with perfectly-reflective materials, designed to highlight the challenge of recovering accurate view-dependent effects. 
Each scene contains $64$ training images and $36$ testing images, all with different random poses. Each training image was rendered with distinct illumination, and the test set was all rendered using the illumination of the reference image from the training set. Please see the supplemental material for additional results and videos.

\newcommand{\appleleft}{0.9in}
\newcommand{\applebottom}{1.1in}
\newcommand{\appleright}{1.3in}
\newcommand{\appletop}{0.68in}

\newcommand{\fireleft}{1.35in}
\newcommand{\firebottom}{0.4in}
\newcommand{\fireright}{1.35in}
\newcommand{\firetop}{0in}

\newcommand{\dollsleft}{0.6in}
\newcommand{\dollsbottom}{1in}
\newcommand{\dollsright}{0.8in}
\newcommand{\dollstop}{1.4in}

\newcommand{\figfivewidth}{0.135\linewidth}

\begin{figure*}[t]
    \centering
    \begin{tabular}{@{}c@{\,\,}c@{\,\,}|@{\,\,}c@{\,\,}c@{\,\,}c@{\,\,}@{\,}c@{\,\,}c@{}}
    {\includegraphics[width=\figfivewidth, trim={1in} {0.7in} {1in} {1.1in}, clip]{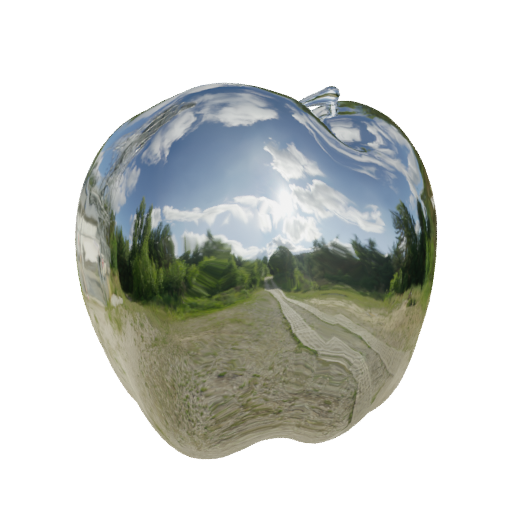}} & 
    {\includegraphics[width=\figfivewidth, trim={0} {0} {0} {0}, clip]{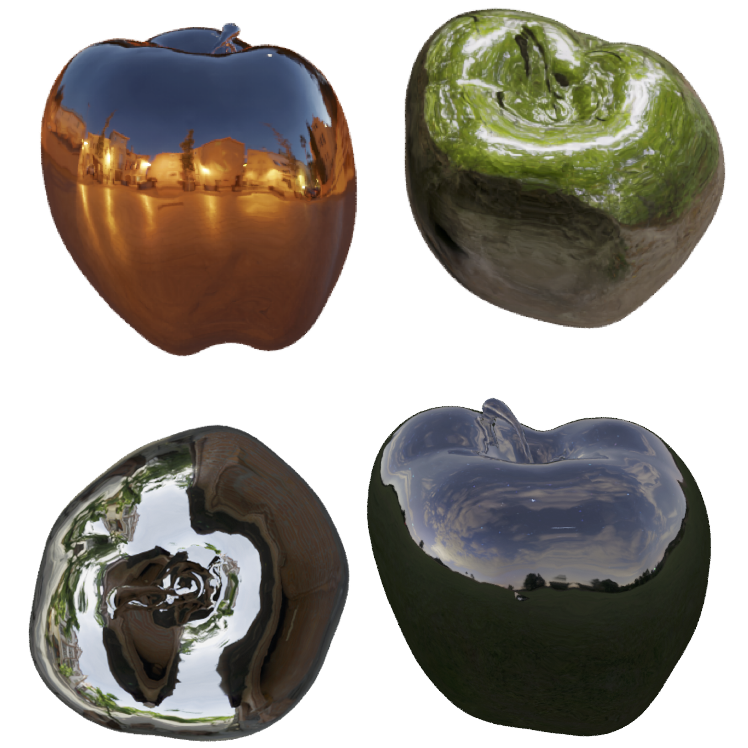}} & 
    {\includegraphics[width=\figfivewidth, trim={\appleleft} {\applebottom} {\appleright} {\appletop}, clip]{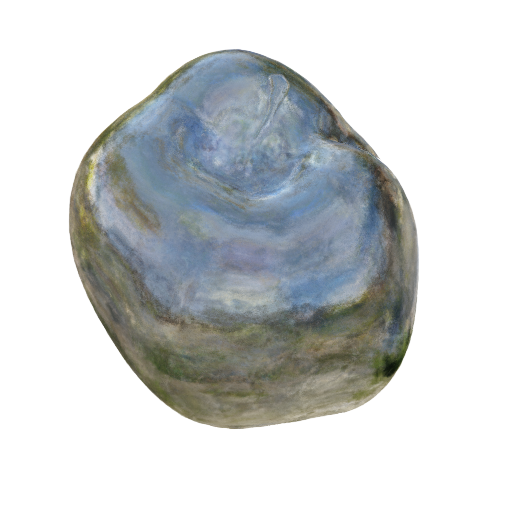}} & 
    {\includegraphics[width=\figfivewidth, trim={\appleleft} {\applebottom} {\appleright} {\appletop}, clip]{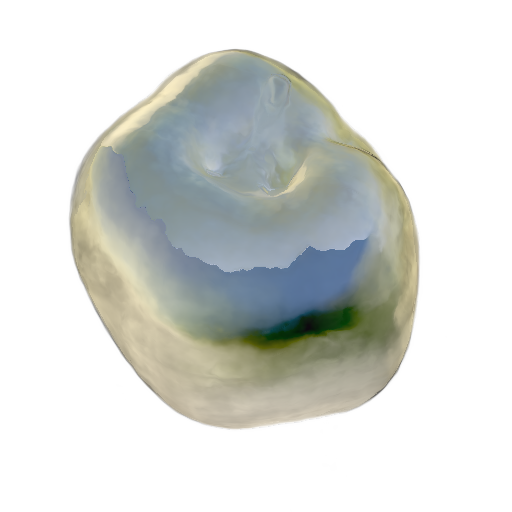}} & 
    {\includegraphics[width=\figfivewidth, trim={\appleleft} {\applebottom} {\appleright} {\appletop}, clip]{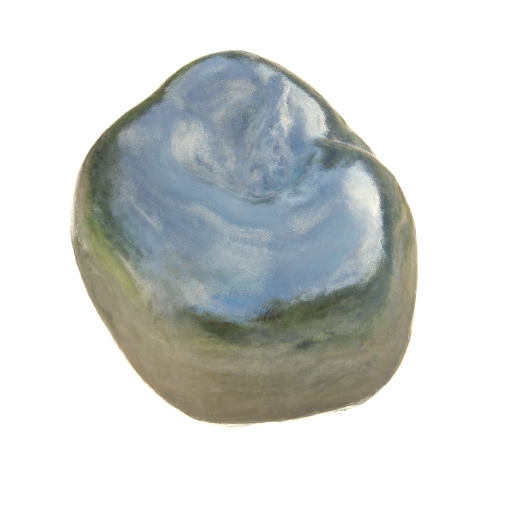}} & 
    {\includegraphics[width=\figfivewidth, trim={\appleleft} {\applebottom} {\appleright} {\appletop}, clip]{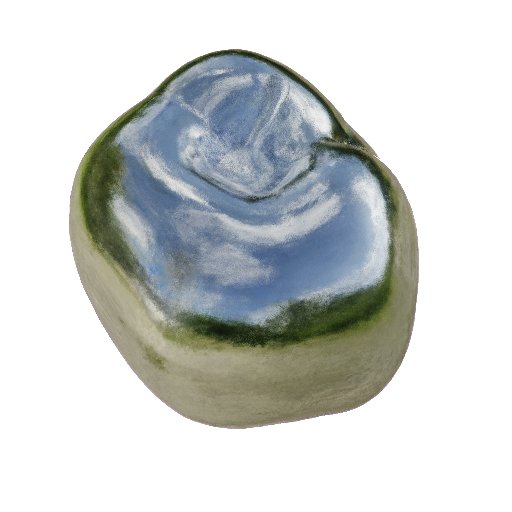}} &
    {\includegraphics[width=\figfivewidth, trim={\appleleft} {\applebottom} {\appleright} {\appletop}, clip]{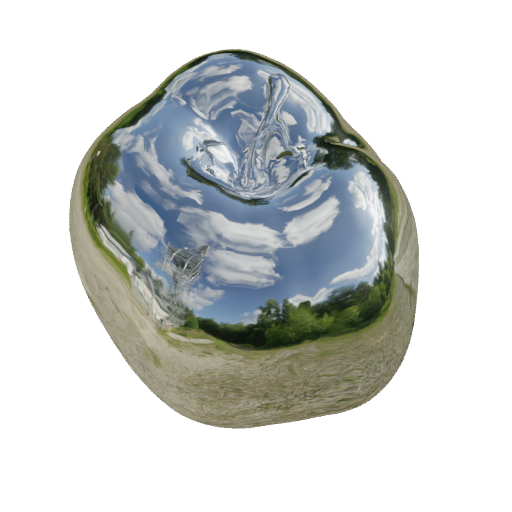}} \\
    {\includegraphics[width=\figfivewidth, trim={\fireleft} {\firebottom} {\fireright} {\firetop}, clip]{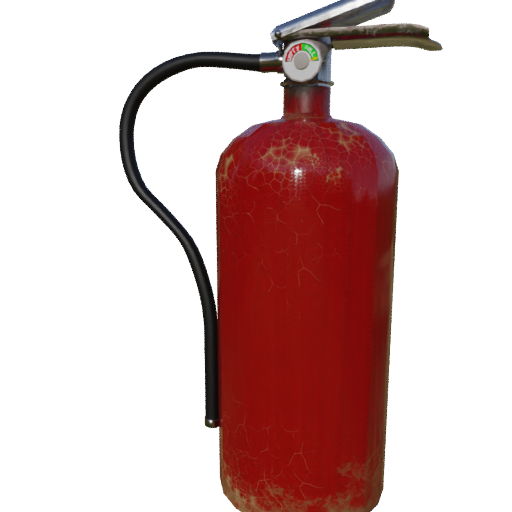}} & 
    {\includegraphics[width=\figfivewidth, trim={0} {0} {0} {0}, clip]{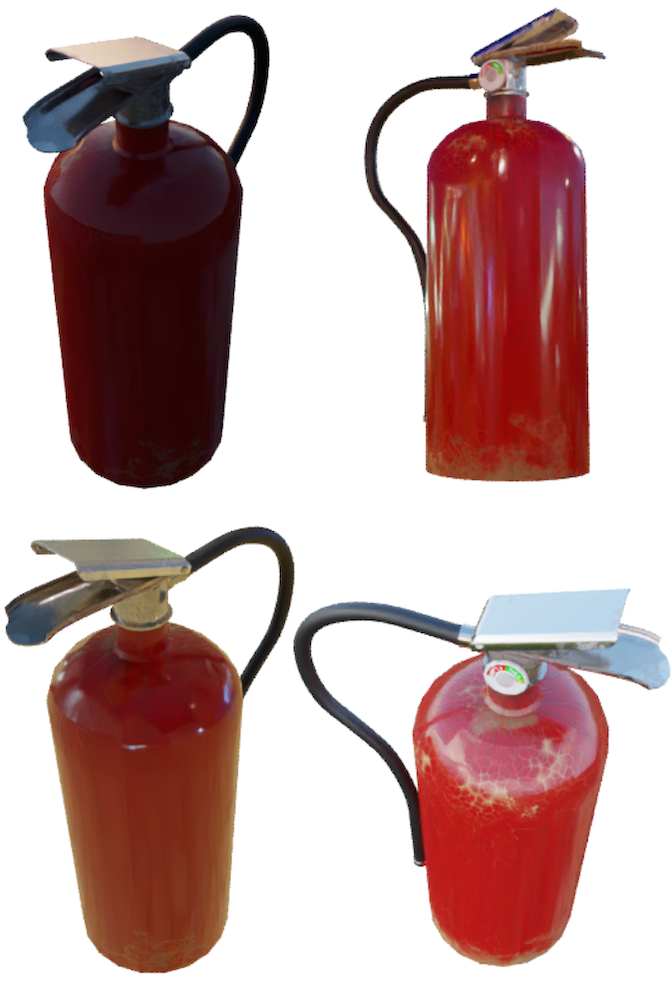}} & 
    {\includegraphics[width=\figfivewidth, trim={\fireleft} {\firebottom} {\fireright} {\firetop}, clip]{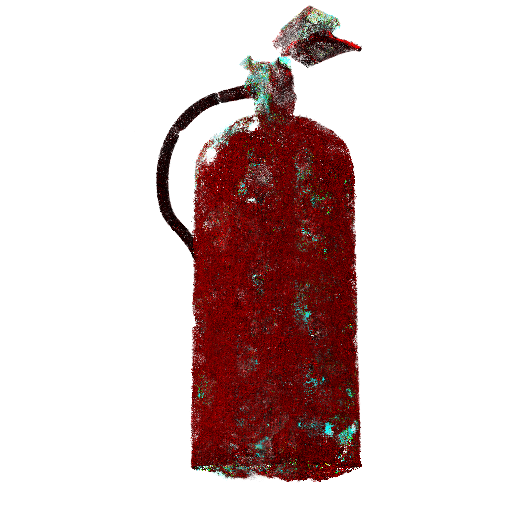}} & 
    {\includegraphics[width=\figfivewidth, trim={\fireleft} {\firebottom} {\fireright} {\firetop}, clip]{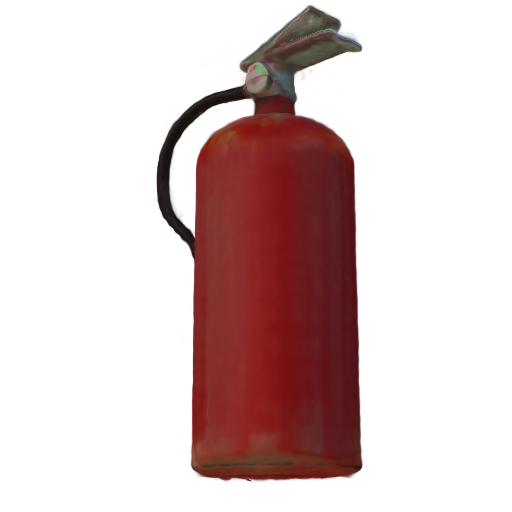}} & 
    {\includegraphics[width=\figfivewidth, trim={\fireleft} {\firebottom} {\fireright} {\firetop}, clip]{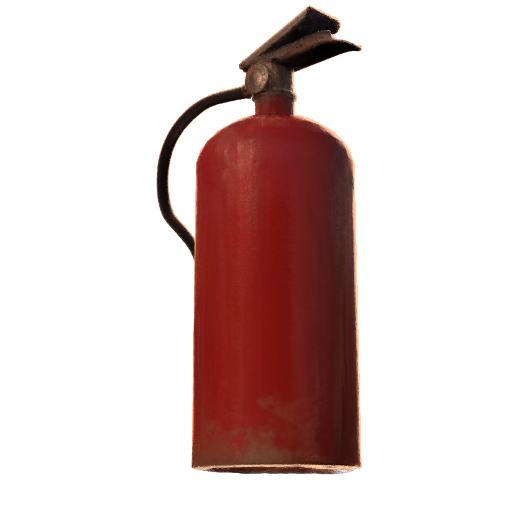} }& 
    {\includegraphics[width=\figfivewidth, trim={\fireleft} {\firebottom} {\fireright} {\firetop}, clip]{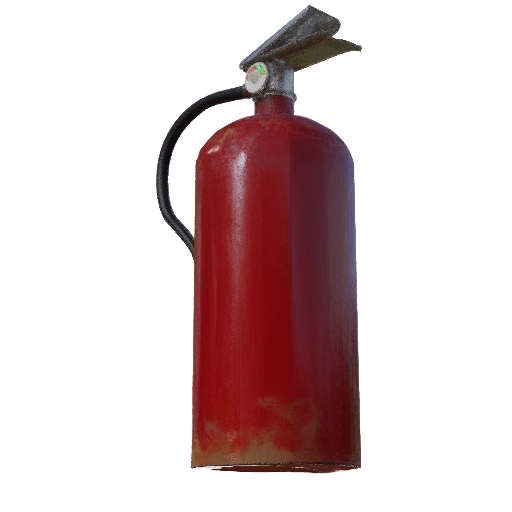}} &
    {\includegraphics[width=\figfivewidth, trim={\fireleft} {\firebottom} {\fireright} {\firetop}, clip]{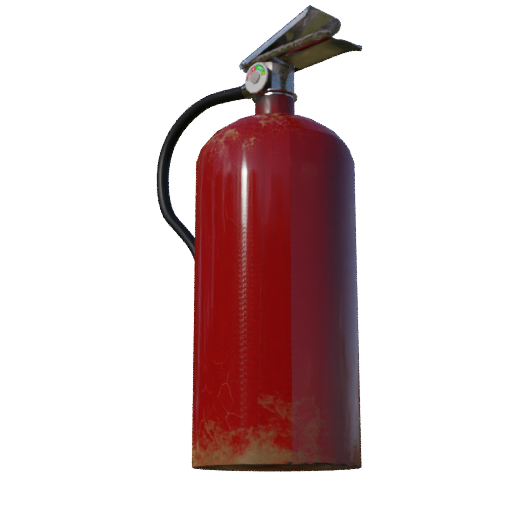}} \\
    {\includegraphics[width=0.115\linewidth, trim={0in} {0in} {0in} {0in}, clip]{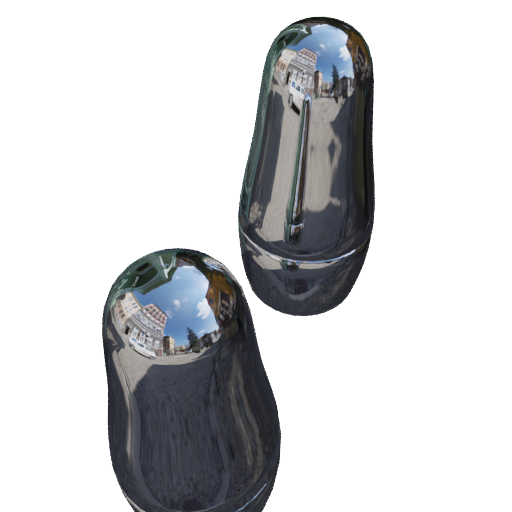}} & 
    {\includegraphics[width=0.11\linewidth, trim={0} {0} {0} {0}, clip]{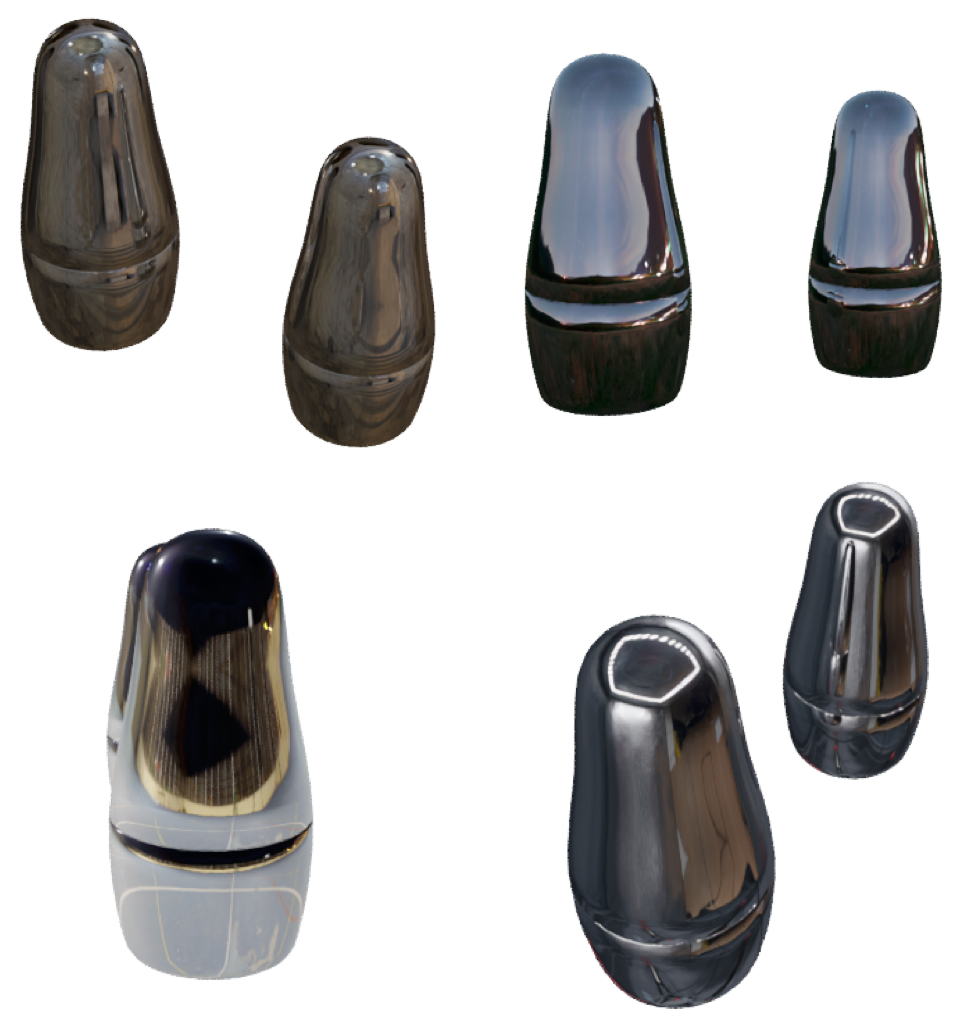}} & 
    {\includegraphics[width=\figfivewidth, trim={\dollsleft} {\dollsbottom} {\dollsright} {\dollstop}, clip]{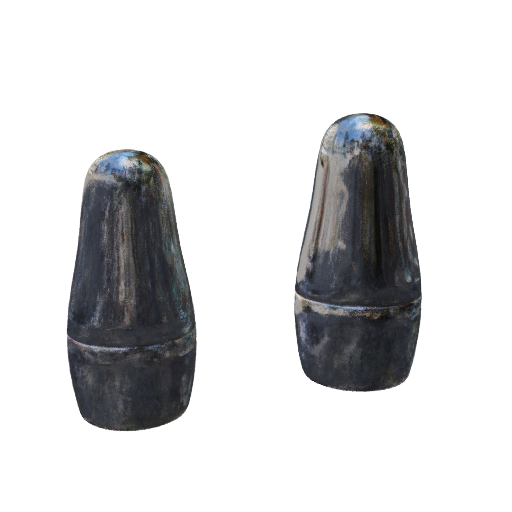}} & 
    {\includegraphics[width=\figfivewidth, trim={\dollsleft} {\dollsbottom} {\dollsright} {\dollstop}, clip]{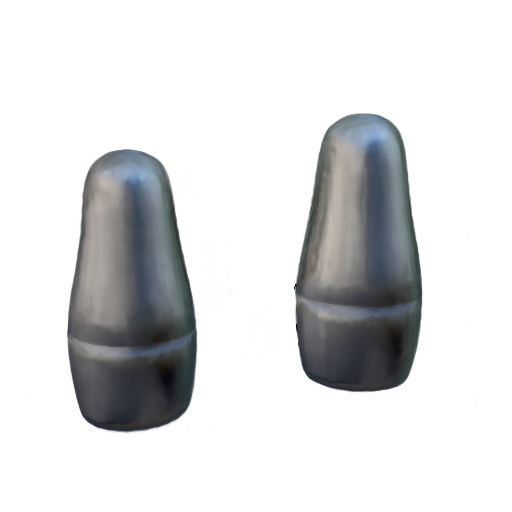}} & 
    {\includegraphics[width=\figfivewidth, trim={\dollsleft} {\dollsbottom} {\dollsright} {\dollstop}, clip]{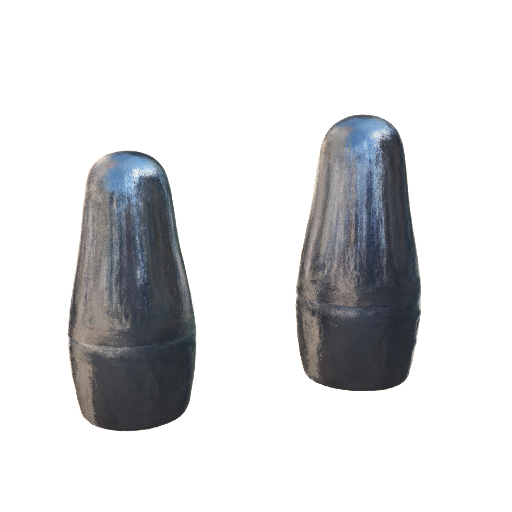}} & 
    {\includegraphics[width=\figfivewidth, trim={\dollsleft} {\dollsbottom} {\dollsright} {\dollstop}, clip]{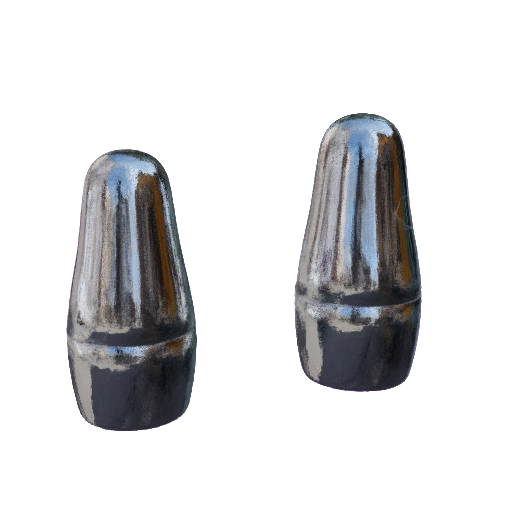}} &
    {\includegraphics[width=\figfivewidth, trim={\dollsleft} {\dollsbottom} {\dollsright} {\dollstop}, clip]{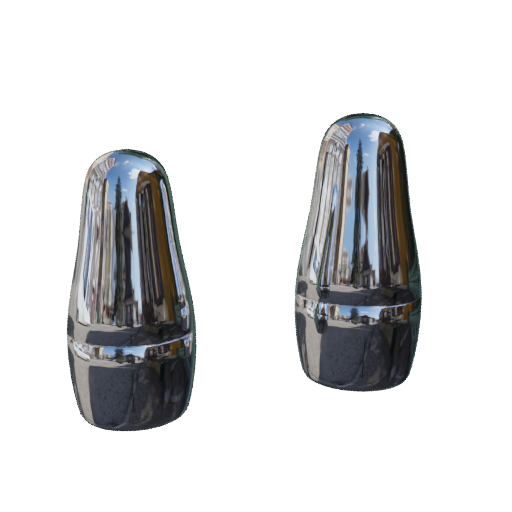}} \\
    \small (a) Reference & \small (b) Sample Inputs & \small (c) NeRFCast + AE & \small (d) NeROIC & \small (e)  IllumiNeRF & \small (f)  Ours  & \small (g)  Ground Truth
    \end{tabular} 
    \caption{\textbf{Visual comparison of novel view renderings on the Objaverse dataset}. 
    (b) We show sample input images under extreme illumination variation. 
    (c) Adding a per-image latent code to NeRF-Casting~\cite{verbin2024nerf} (``NeRFCast + AE'') cannot accurately explain away the variations, leading to erroneous reconstruction.
    (d) Due to the ill-posed nature of the problem, inverse rendering-based methods such as NeROIC~\cite{neroic2022kuang} tend to produce lower-quality renderings with mostly diffuse appearance. 
    (e) IllumiNeRF~\cite{zhao2024illuminerf} leverages diffusion prior for single-image relighting but produces inconsistent output samples, resulting in excessive blur in rendered novel views. Note that IllumiNeRF requires access to the target illumination's environment map as input. We provide IllumiNeRF with the ground truth environment map corresponding to the reference image (a), while other methods only have access to the reference image itself.
    (f) Our method renders accurate appearance with specular highlights close to those in the ground truth images (g). 
    }
    \label{fig:objaverse}
\end{figure*} 

\subsection{Comparisons to State-of-the-Art Methods}

We compare our approach with two methods designed for 3D reconstruction ``in the wild'': NeROIC~\cite{neroic2022kuang} and NeRF-Casting~\cite{verbin2024nerf} with standard appearance embeddings for each image (which we call ``NeRF-Casting + AE'' or ``NeRFCast + AE'' for short).
We also compare our approach with the recent diffusion-based relighting method IllumiNeRF~\cite{zhao2024illuminerf}. IllumiNeRF requires a target environment map for relighting
and we provide it with the ground truth illumination, which is a significant advantage over other techniques as the full environment map contains more information than the reference image. For methods with appearance embeddings (ours and NeRF-Casting + AE), we render novel views using the same embedding as the reference image. Similarly, when rendering test images using NeROIC, we use the estimated illumination of the reference image.

\topic{Synthetic Results}

In Figure~\ref{fig:objaverse}, we qualitatively compare our method's novel view synthesis results with those of prior state-of-the-art approaches. Our results appear significantly more accurate, and all baselines struggle to recover specular highlights. This is particularly apparent in the first and third rows, demonstrating recovery of fine details in the reflections of perfectly-specular objects. The quantitative results in Table~\ref{tab:standard_eval} show that our method significantly outperforms all baselines on both splits of the dataset, even when we provide the IllumiNeRF~\cite{zhao2024illuminerf} baseline with true environment maps corresponding to the reference image.

\begin{table}[h!]
\centering
\resizebox{\columnwidth}{!}{%
\begin{tabular}{@{}l@{\,\,}|@{\,\,}c@{\,\,}c@{\,\,}c|c@{\,\,}c@{\,\,}c@{}}
\toprule
& \multicolumn{3}{c}{Standard Assets} & \multicolumn{3}{|c}{Shiny Assets} \\
\midrule
\textbf{Method} & \textbf{PSNR$\uparrow$} & \textbf{SSIM$\uparrow$} & \textbf{LPIPS$\downarrow$}  & \textbf{PSNR$\uparrow$} & \textbf{SSIM$\uparrow$} & \textbf{LPIPS$\downarrow$}\\
\midrule
NeROIC ~\cite{neroic2022kuang} & 26.13 & 0.935 & 0.088  & 22.14 & 0.880 & 0.113\\
NeRFCast~\cite{verbin2024nerf} + AE & 27.53 &0.941 &	0.067 & 21.80 & 0.874 & 0.108\\
IllumiNeRF~\cite{zhao2024illuminerf} w/ known light & 29.22 & 0.958 & 0.057 & 23.46 & 0.881 & 0.095 \\
Ours & \textbf{31.34} & \textbf{0.966} &\textbf{0.053} & \textbf{26.54} & \textbf{0.911} &	\textbf{0.090}\\
\bottomrule
\end{tabular}
}
\vspace{-0.05in}
\caption{
\textbf{Novel view relighting on Objaverse.} Our approach of relighting and robustly optimizing a radiance field outperforms previous state-of-the-art methods in rendering novel views under target lighting. We outperform prior work despite giving the baseline a significant advantage by providing it the true lighting.
}
\label{tab:standard_eval}
\end{table}

\newcommand{\keyleft}{1.15in}
\newcommand{\keybottom}{1.4in}
\newcommand{\keyright}{2.125in}
\newcommand{\keytop}{2.6in}

\newcommand{\rcleft}{1.75in}
\newcommand{\rcbottom}{1.5in}
\newcommand{\rcright}{1.35in}
\newcommand{\rctop}{1.3in}

\newcommand{\bunnyleft}{2in}
\newcommand{\bunnybottom}{1.6in}
\newcommand{\bunnyright}{1.4in}
\newcommand{\bunnytop}{0.75in}

\newcommand{\colorleft}{1.65in}
\newcommand{\colorbottom}{1.15in}
\newcommand{\colorright}{1.3in}
\newcommand{\colortop}{1.3in}

\newcommand{\naviwidth}{0.158\linewidth}


\begin{figure*}[t]
    \centering
    \begin{tabular}{@{}c@{\,\,}c@{\,\,}|@{\,\,}c@{\,\,}c@{\,\,}c@{\,\,}@{\,}c@{\,\,}c@{}}
    {\includegraphics[width=0.13\linewidth, trim={2.15in} {1.8in} {1.75in} {1.3in}, clip]{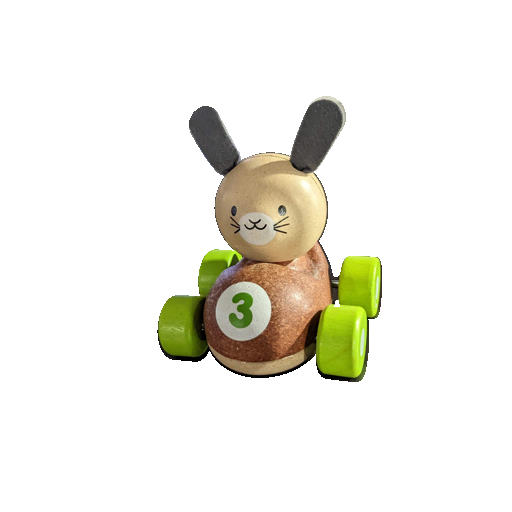}} & 
    {\includegraphics[width=0.16\linewidth, trim={0} {0} {0} {0}, clip]{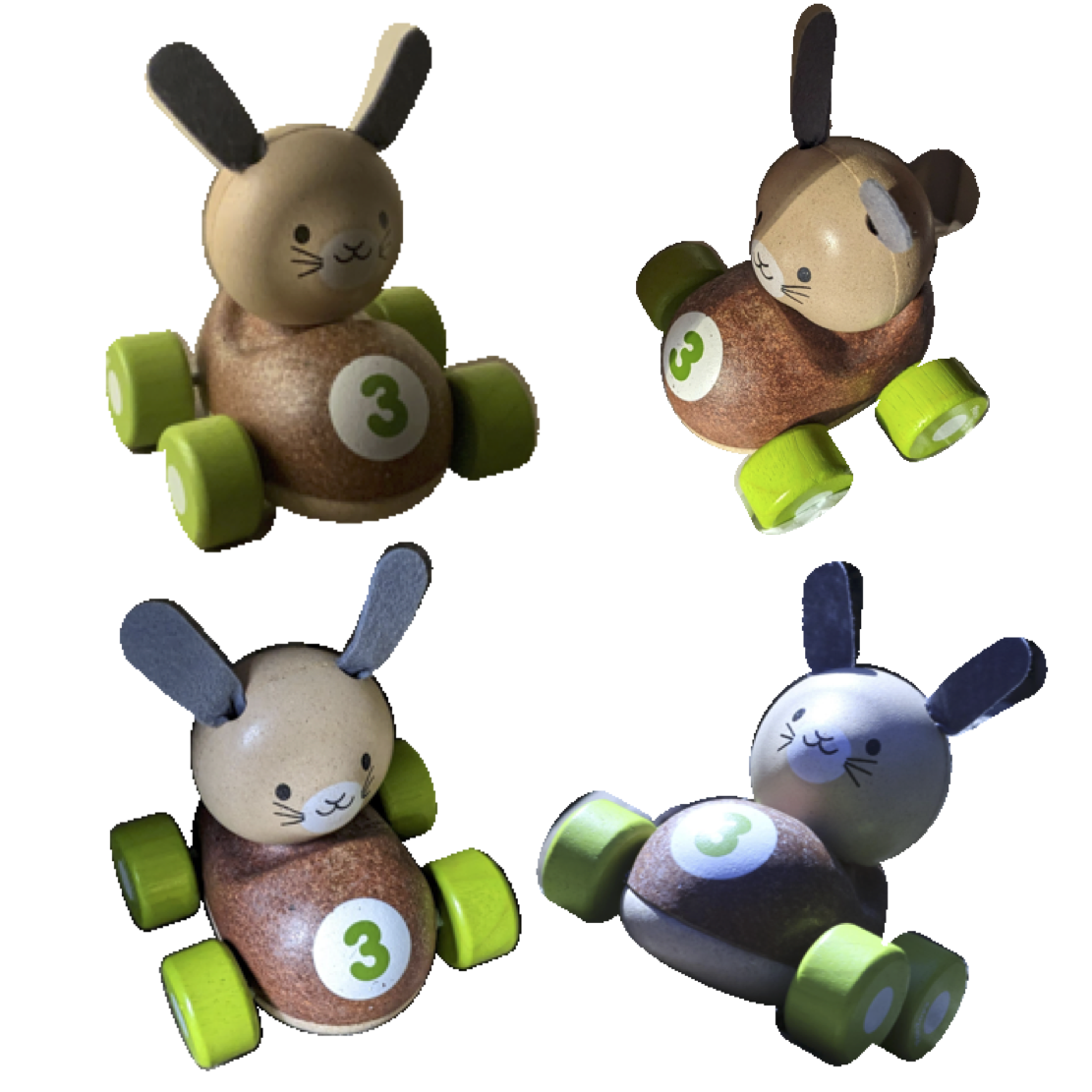}} & 
    {\includegraphics[width=0.13\linewidth, trim={\bunnyleft} {\bunnybottom} {\bunnyright} {\bunnytop}, clip]{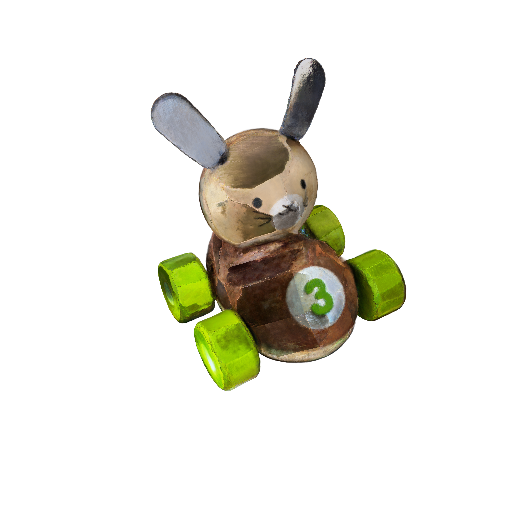}} & 
    {\includegraphics[width=0.13\linewidth, trim={\bunnyleft} {\bunnybottom} {\bunnyright} {\bunnytop}, clip]{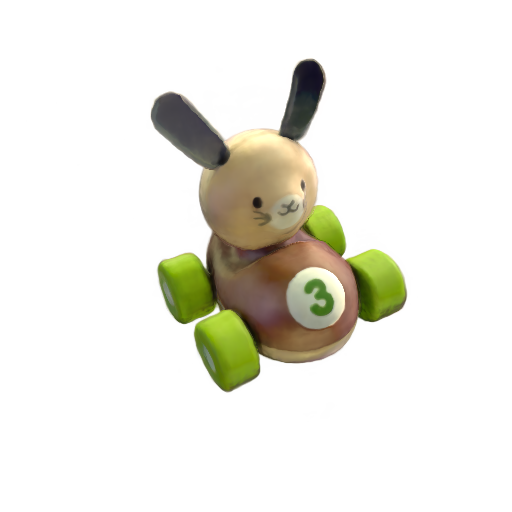}} & 
    {\includegraphics[width=0.13\linewidth, trim={\bunnyleft} {\bunnybottom} {\bunnyright} {\bunnytop}, clip]{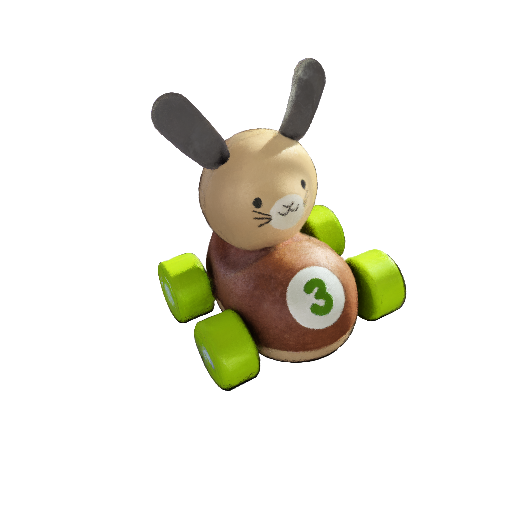}} &
    {\includegraphics[width=0.13\linewidth, trim={\bunnyleft} {\bunnybottom} {\bunnyright} {\bunnytop}, clip]{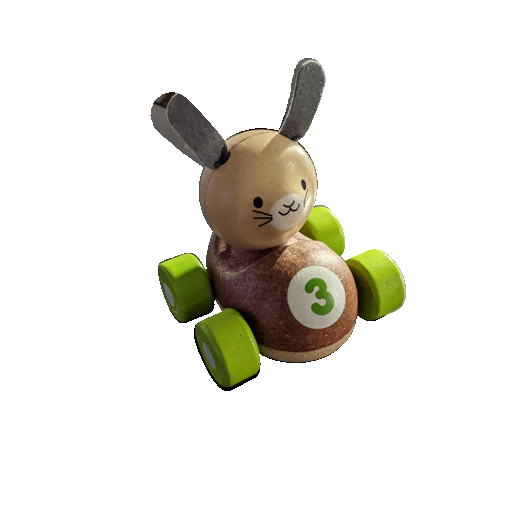}} \\
    {\includegraphics[width=\naviwidth, trim={1.4in} {1.05in} {0.8in} {1.2in}, clip]{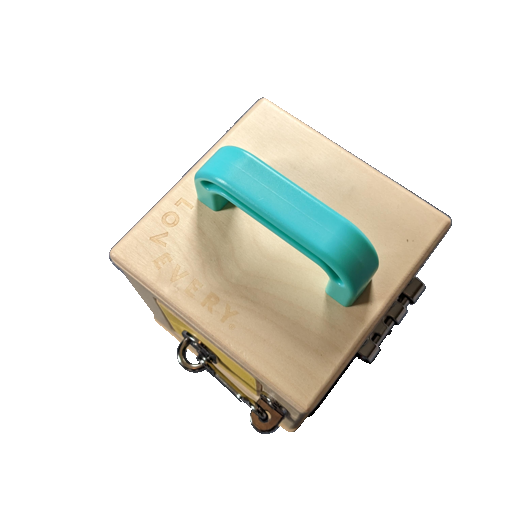}} & 
    {\includegraphics[width=\naviwidth, trim={0} {0} {0} {0}, clip]{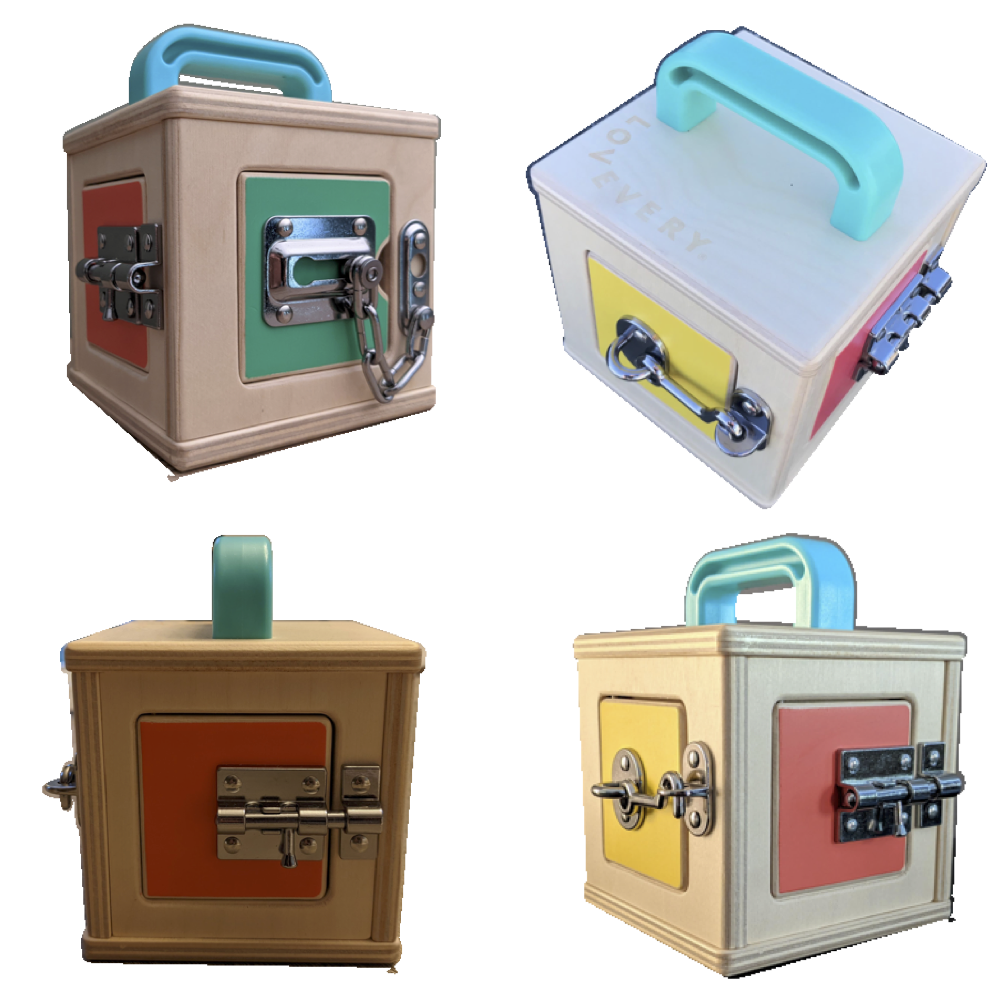}} & 
    {\includegraphics[width=.14\linewidth, trim={\colorleft} {\colorbottom} {\colorright} {\colortop}, clip]{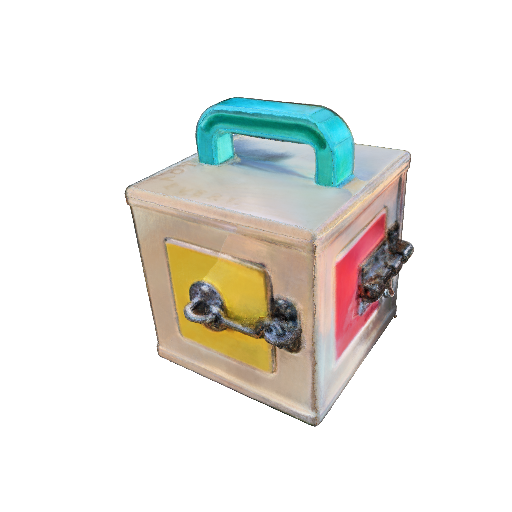}} & 
    {\includegraphics[width=.14\linewidth, trim={\colorleft} {\colorbottom} {\colorright} {\colortop}, clip]{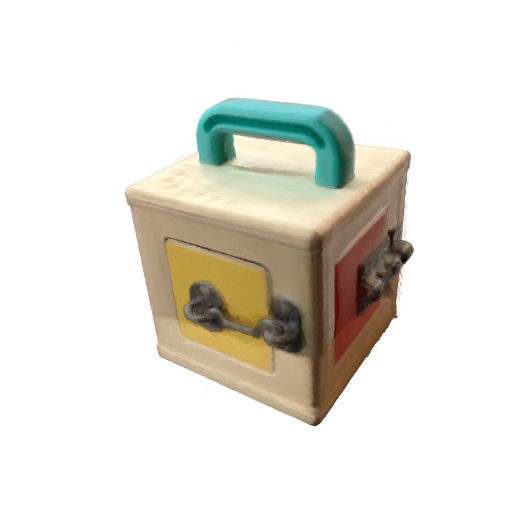}} & 
    {\includegraphics[width=.14\linewidth, trim={\colorleft} {\colorbottom} {\colorright} {\colortop}, clip]{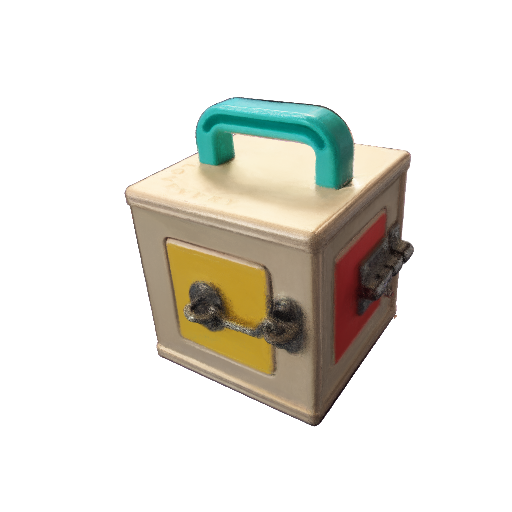}} &
    {\includegraphics[width=0.14\linewidth, trim={\colorleft} {\colorbottom} {\colorright} {\colortop}, clip]{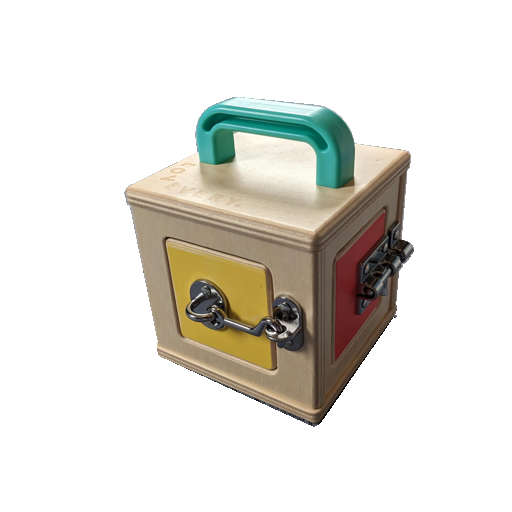}} \\
    \includegraphics[width=\naviwidth, trim={2.1in} {1.7in} {2in} {2.6in}, clip]{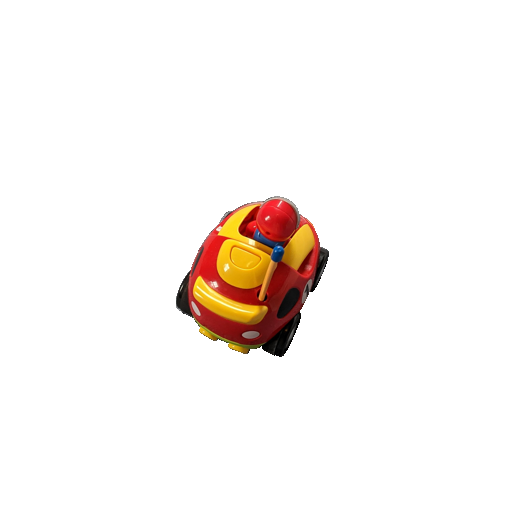} & 
    \includegraphics[width=\naviwidth, trim={0} {0} {0} {0}, clip]{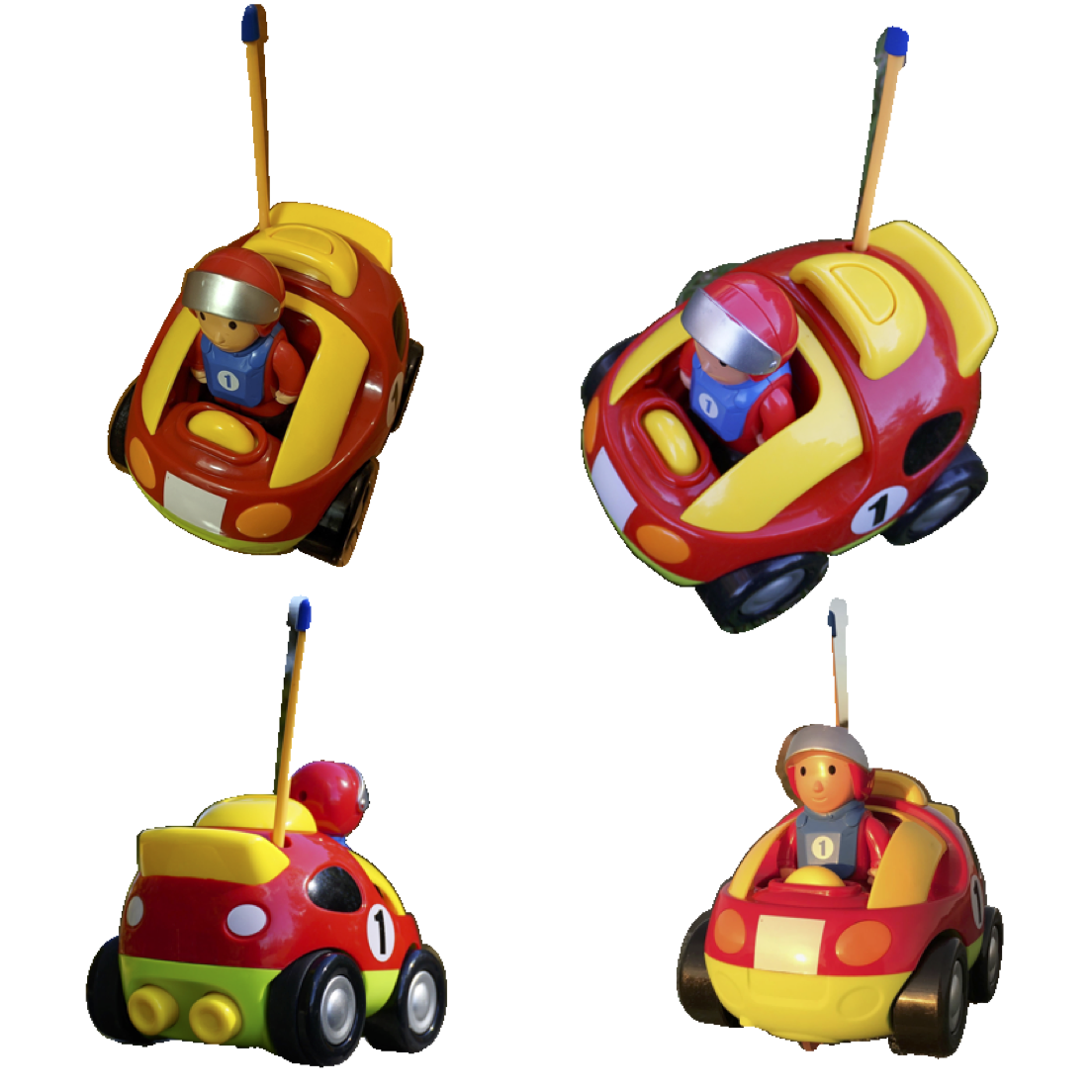} & 
    \includegraphics[width=\naviwidth, trim={\rcleft} {\rcbottom} {\rcright} {\rctop}, clip]{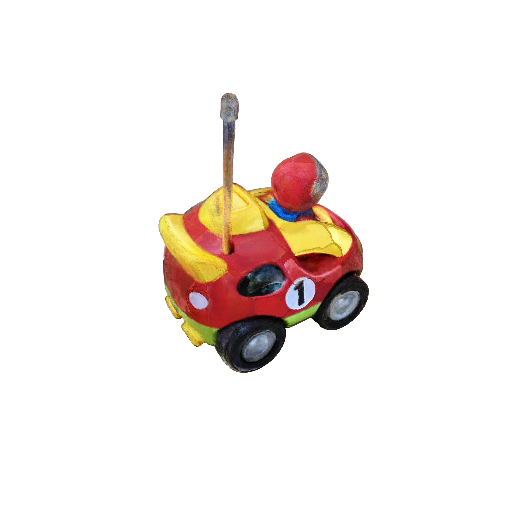} & 
    \includegraphics[width=\naviwidth, trim={\rcleft} {\rcbottom} {\rcright} {\rctop}, clip]{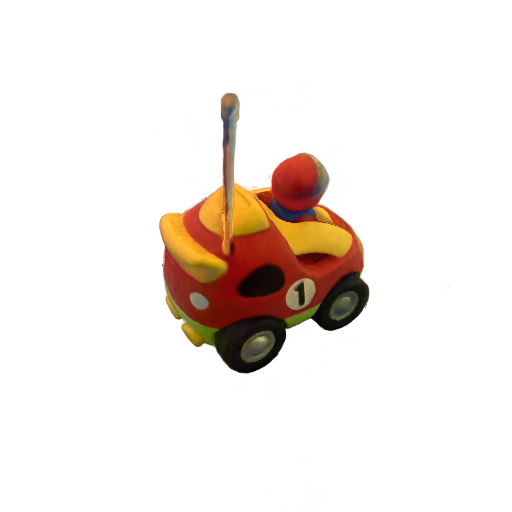} & 
    \includegraphics[width=\naviwidth, trim={\rcleft} {\rcbottom} {\rcright} {\rctop}, clip]{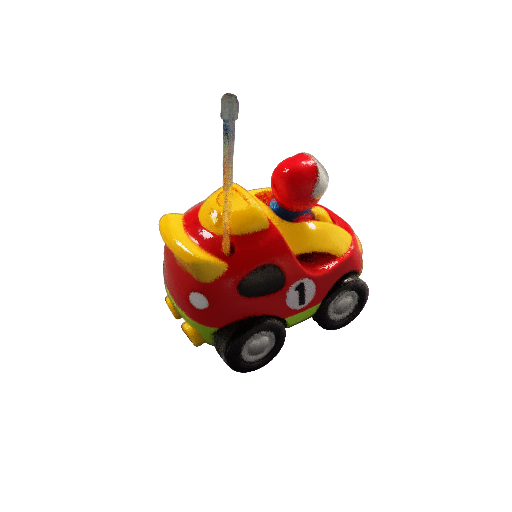} &
    \includegraphics[width=\naviwidth, trim={\rcleft} {\rcbottom} {\rcright} {\rctop}, clip]{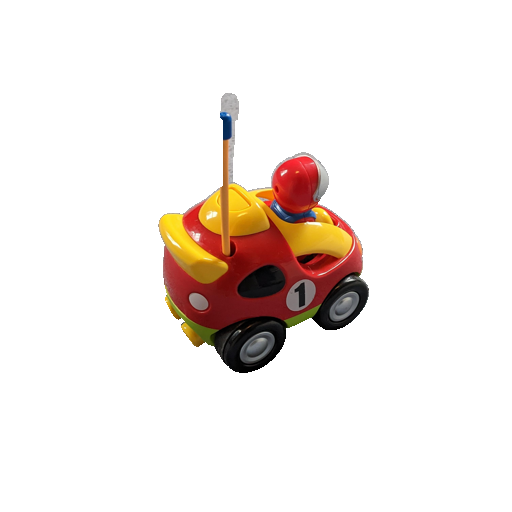} \\
    \small (a) Reference & \small (b) Input Samples & \small (c) NeRFCast + AE & \small (d) NeROIC  & \small (e)  Ours  & \small (f)  Ground Truth
    \end{tabular} 
    \caption{\textbf{Comparison on real world photos.} We use our method to reconstruct objects from in-the-wild photos taken in different environments. Our method can render novel views under the illumination conditions of any input image we select as the reference. Unlike prior work, our technique accurately preserves shadows (\eg the bunny's ear shadow in the first row) and reflections (\eg the box's handle in the second row and the car's specularities in the last row) that appear in the reference image we would like to match. 
    }
    \label{fig:navi}
\end{figure*}

\topic{Real Photographs}
The top row of Figure~\ref{fig:teaser} shows photographs we captured of a chrome-plated figurine in different environments. The resulting synthesized novel views show that our method can render accurate images corresponding to the selected reference illumination. 
In the bottom of Figure~\ref{fig:teaser} and in Figure~\ref{fig:navi} we show reconstruction results on captured photographs from NAVI. The results show that our method can preserve shadows and specular highlights. Since NAVI also provides multiview data for each illumination, we can quantiatively evaluate the novel view synthesis performance on held-out views under the same illumination as the reference. We evaluate the method on all of the $18$ NAVI scenes that have at least $64$ frames, and we show that our method also achieves the best quantitative results across all prior methods, as shown in Table~\ref{tab:navi}. Since NAVI does not provide ground truth environment maps, we could not evaluate IllumiNeRF on this dataset.

\begin{table}[h!]
\centering
\small
\begin{tabular}{@{}l|ccc@{}}
\toprule
\textbf{Method} & \textbf{PSNR$\uparrow$} & \textbf{SSIM$\uparrow$} & \textbf{LPIPS$\downarrow$} \\
\midrule
NeRF-Casting~\cite{verbin2024nerf} + AE  &  22.67 & 0.906 & 0.074 \\
NeROIC \cite{neroic2022kuang} & 24.01 & 0.918 & 0.079 \\
Our Model & \textbf{25.55} & \textbf{0.929} & \textbf{0.060} \\
\bottomrule
\end{tabular}
\vspace{-0.05in}
\caption{
\textbf{Novel view relighting on real photos from NAVI.}
 We show that we outperform prior work on synthesizing novel views from ``in-the-wild'' images.
}
\label{tab:navi}
\end{table}

\subsection{Ablation Studies}
Here, we provide ablations designed to test the different components of our model.

\topic{Number of Frames} A major benefit of our approach is that it jointly relights all frames. Table~\ref{tab:frames_ablation} shows our view synthesis results when relighting the images using different numbers of simultaneously-predicted frames $N$. Increasing the number of predicted frames improves 3D reconstruction, hence our choice for predicting $N=64$ images.

\begin{table}[h!]
\centering
\resizebox{\columnwidth}{!}{%
\begin{tabular}{@{}l|c@{\,\,}c@{\,\,}c|c@{\,\,}c@{\,\,}c@{}}
\toprule
& \multicolumn{3}{c}{Standard Assets} & \multicolumn{3}{|c}{Shiny Assets} \\
\midrule
\textbf{\# frames $N$} & \textbf{PSNR$\uparrow$} & \textbf{SSIM$\uparrow$} & \textbf{LPIPS$\downarrow$}  & \textbf{PSNR$\uparrow$} & \textbf{SSIM$\uparrow$} & \textbf{LPIPS$\downarrow$}\\
\midrule
1 frame & 28.39 & 0.943 & 0.079 & 23.85 & 0.889 & 0.105\\
8 frames & 30.43 & 0.956 &  0.067 & 25.62 & 0.901 & 0.095 \\
16 frames  & 30.49 & 0.958 & 0.063 & 25.52 & 0.900 & 0.097 \\
32 frames & 30.72 & 0.961 & 0.060 & 26.14 & 0.908 & 0.090\\
64 frames (ours) & \textbf{31.34} & \textbf{0.966} & \textbf{0.053} & \textbf{26.54} & \textbf{0.911} & \textbf{0.090} \\
\bottomrule
\end{tabular}
}
\vspace{-0.05in}
\caption{
\textbf{The effect of the number of frames $N$ output simultaneously by our model.} Increasing the number of frames simultaneously processed by our model improves its performance, therefore we use $N=64$.}
\label{tab:frames_ablation}
\end{table}

\topic{Dataset} Next, we demonstrate the importance of including purely-specular objects in our training data. To do this, we train two 16-frames version of our model for $70$K steps, with and without the purely-specular data augmentation described in Section~\ref{sec:training}. Table~\ref{tab:dataset_ablation} shows that including the specular materials in the training set improves performance on specular objects and, perhaps surprisingly, that it is also beneficial for our general less-specular (but not perfectly-diffuse) dataset. 

\begin{table}[h!]
\centering
\resizebox{\columnwidth}{!}{%
\begin{tabular}{@{}l|c@{\,\,}c@{\,\,}c|c@{\,\,}c@{\,\,}c@{}}
\toprule
& \multicolumn{3}{c}{Standard Assets} & \multicolumn{3}{|c}{Shiny Assets} \\
\midrule
\textbf{Training data} & \textbf{PSNR$\uparrow$} & \textbf{SSIM$\uparrow$} & \textbf{LPIPS$\downarrow$}  & \textbf{PSNR$\uparrow$} & \textbf{SSIM$\uparrow$} & \textbf{LPIPS$\downarrow$}\\
\midrule
Standard assets & 29.16 & 0.938 & 0.066   & 22.34 & 0.874 & 0.113  \\
Standard + shiny assets &   \textbf{29.26} & \textbf{0.939} & \textbf{0.066} &  \textbf{22.91} & \textbf{0.877} &  \textbf{0.111} \\
\bottomrule
\end{tabular}
}
\vspace{-0.05in}
\caption{
\textbf{Training dataset ablation.}
We show that training on assets with standard materials as well as objects with highly reflective materials improves performance for both types of assets. 
}
\label{tab:dataset_ablation}
\end{table}

\topic{Shading Embeddings} Table~\ref{tab:nerf_ablation} demonstrates the effect of our shading embeddings on the 3D reconstruction of the shiny synthetic scenes harmonized by our relighting model. 
As explained in Section~\ref{sec:nerf}, the shading embeddings allow the model to slightly shift the normal vectors used for computing appearance separately for each image, which enables absorbing small inconsistencies in the relit images. This improves upon the same NeRF model with no per-image embeddings (``No embeddings''). In contrast, using standard embeddings as in NeRF-W~\cite{martinbrualla2020nerfw} (``Shading embeddings'') has an adverse effect: the per-image codes allow the model to encode specularities as per-image diffuse color, which results in unrealistic and poor reconstructions.

\begin{table}[h!]
\centering
\small
\begin{tabular}{@{}l|ccc@{}}
\toprule
\textbf{Method} & \textbf{PSNR}$\uparrow$ & \textbf{SSIM }$\uparrow$ & \textbf{LPIPS}$\downarrow$ \\
\midrule
No embeddings  & 26.02 & 0.907 & 0.094 \\
Appearance embeddings & 24.38 & 0.896 & 0.096 \\
Shading embeddings (ours) &\textbf{26.54} & \textbf{0.911} &	\textbf{0.090} \\
\bottomrule
\end{tabular}
\caption{
\textbf{Per-image embedding ablation.} We show that our shading embeddings improve performance compareo NeRF-Casting without any per-image code and that standard appearance embeddings have an adverse effect on reconstruction quality.
}
\label{tab:nerf_ablation}
\end{table}

\vspace{-2mm}
\section{Discussions and Conclusions}

\topic{Limitations}
Our approach requires object masks and accurate camera poses as inputs. Camera poses, in particular, can be challenging to compute from images of highly reflective objects that do not have reliable features for matching, but are required by most 3D reconstruction methods.
However, there has been significant recent progress in camera pose estimation using learning-based techniques~\cite{dust3r_cvpr24,zhang2024cameras,brachmann2024scene,monst3r} which do not rely on explicit feature matching.
We believe that leveraging strong generative priors for joint pose estimation and 3D reconstruction is an exciting direction for future research.

\topic{Conclusions}
In this paper, we address the challenge of reconstructing 3D objects from images captured under extreme illumination variation. 
Our core insight is to simultaneously relight all input images to match the illumination of a chosen reference image, thereby harmonizing lighting conditions among input images.
We achieve this by training a diffusion-based multiview relighting model. 
Using these relit images, we apply a 3D reconstruction model that is robust to residual inconsistencies, enabling accurate reconstruction of the object's shape and view-dependent appearance. 
We demonstrate that our method effectively reconstructs objects with complex shapes and materials from images taken under drastically different lighting conditions.
Our work highlights a promising direction for leveraging strong generative priors to tackle the inherently ill-posed problems of 3D reconstruction. 

\topic{Acknowledgements} We would like to thank Matthew Burruss and Xiaoming Zhao for their help with the rendering pipeline. We also thank Ben Poole, Alex Trevithick, Stan Szymanowicz, Rundi Wu, David Charatan, Jiapeng Tang, Matthew Levine, Ruiqi Gao, Ricardo Martin-Brualla, and Aleksander Hołyński for fruitful discussions.

{
    \small
    \bibliographystyle{ieeenat_fullname}
    \bibliography{main}
}

\appendix
\setcounter{equation}{0}
\setcounter{table}{0}
\setcounter{figure}{0}
\def\theequation{S\arabic{equation}}
\def\thetable{S\arabic{table}}
\def\thefigure{S\arabic{figure}}

\clearpage
\setcounter{page}{1}
\maketitlesupplementary




\begin{figure}
\begin{center}
\centering

\includegraphics[width=1.0\linewidth, trim=0 0 0 0, clip]{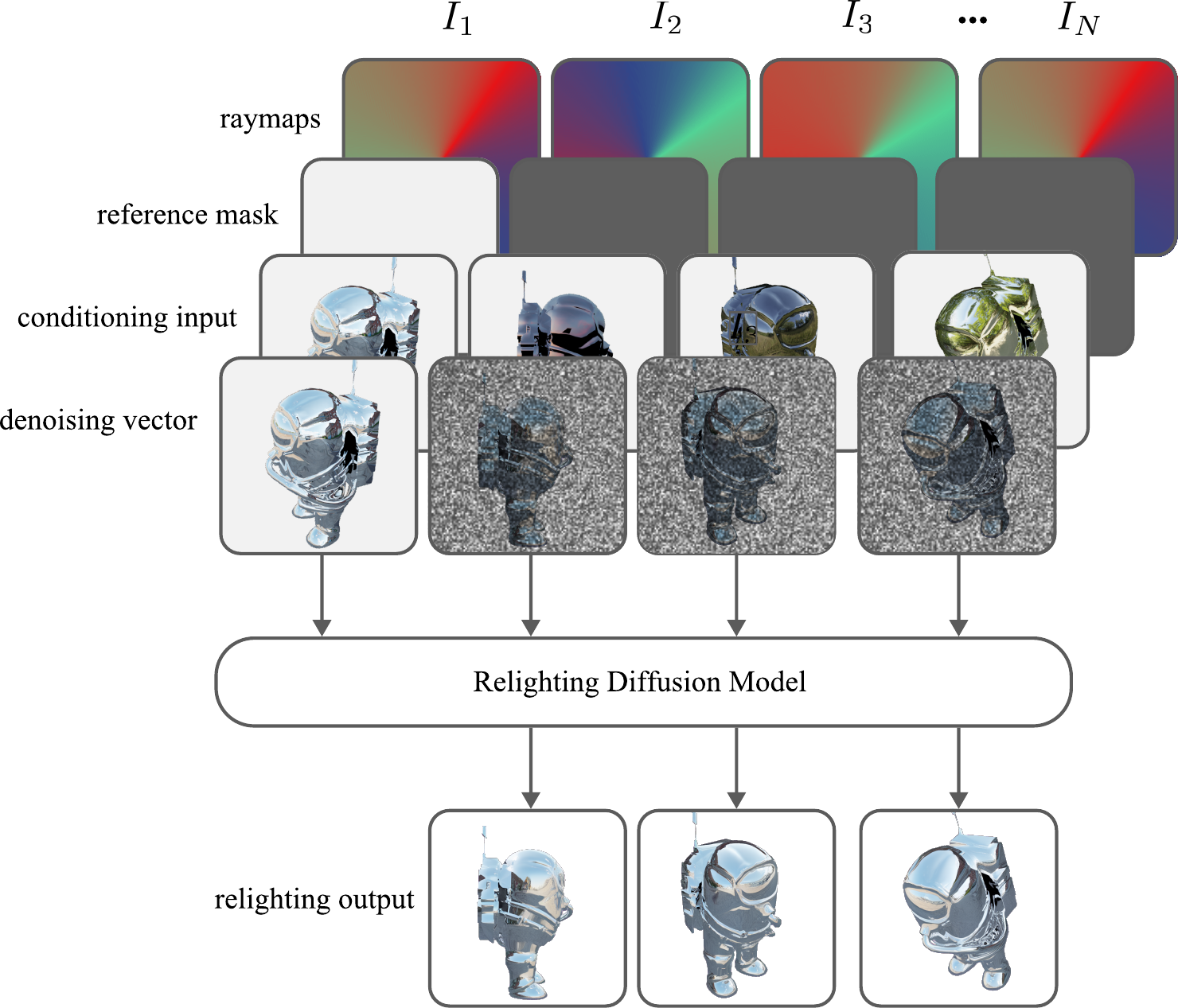}

\captionof{figure}{
\textbf{Relighting model overview.} For each input image, we pass to the model raymaps containing the pose information, a binary mask to highlight which frame is being treated as the reference, and the conditioning images of the object under varying illumination. We concatenate all the inputs to the image being currently denoised. \textit{Note that both the conditioning inputs and target images are the latent encoding of the images, but we show them as images here for simplicity.}
}
\label{fig:diffusion_overview}
\end{center}
\end{figure}

\section{Supplemental Webpage and Videos}
Please refer to our webpage \url{https://relight-to-reconstruct.github.io} to view our video reconstruction results and baseline comparisons on our full synthetic dataset, as well as examples from NAVI. We also show output examples from our relighting model and compare the consistency of our relightings and recent state-of-the-art generative relighting from IllumiNeRF~\cite{zhao2024illuminerf}.

\section{Diffusion Training and Sampling}
In Figure~\ref{fig:diffusion_overview} we show the detailed inputs of our diffusion model. Although our diffusion model operates in the latent space (so the conditioning and targets are both the latents of the encoded images), we refer to them as images for clarity. The raymaps consist of the ray origins, ray directions, and focal length associated with the image for each pixel. We downscale them from $512\times512$ to $64\times64$ to match the dimensions of our latents. The reference mask is a $64 \times 64$ binary mask that is $1$ for the reference image, and $0$ for the images we are relighting. For each denoising step during training and inference, we also pass the latents original images that we want to relight. Note that we do not denoise the reference image, and pass a clean copy of it to the model. While this design redundantly passes the reference image twice, this is needed since the model parameters are shared for all the images. 

We train the relighting diffusion model using a DDPM schedule, with beta values that start at $8.5\cdot 10^{-4}$ and end at $1.2\cdot 10^{-2}$ increasing linearly over $1024$ steps. For objective, we use velocity prediction. During inference, we use DDIM sampling with 50 inference steps.
We use a learning rate of $10^{-4}$, with $10$K warm-up steps. Note that we reset the learning rate schedule each time we fine-tune the model to relight larger number of frames.

\section{Shading embeddings visualization}
\label{sec:shading}

\begin{figure}[t]
    \centering
    {
    \setlength\arrayrulewidth{1pt}
    \begin{tabular}{@{}c@{\,\,}c@{\,\,}c@{}}
    \includegraphics[width=0.49\linewidth, trim={80px} {50px} {65px} {50px}, clip]{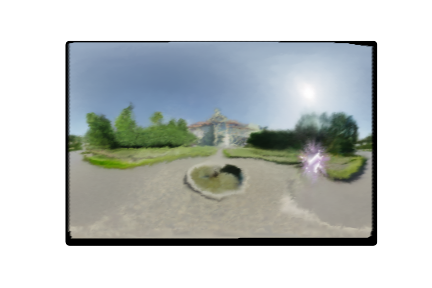} & 
    \includegraphics[width=0.49\linewidth, trim={80px} {50px} {65px} {50px}, clip]{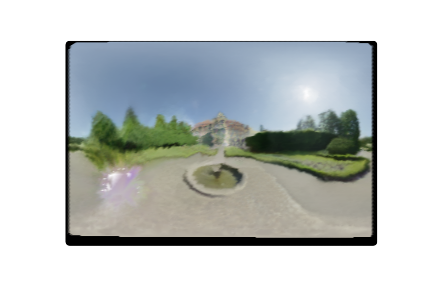} \\
    \includegraphics[width=0.49\linewidth, trim={80px} {50px} {65px} {50px}, clip]{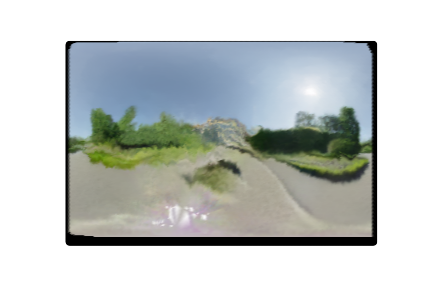} & 
    \includegraphics[width=0.49\linewidth, trim={80px} {50px} {65px} {50px}, clip]{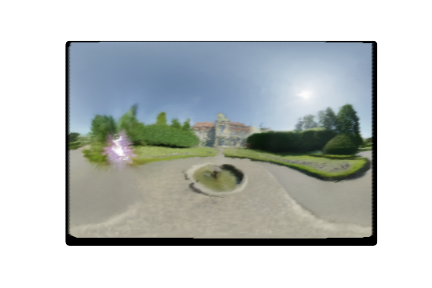} \\
    \end{tabular} 
    }
    \caption{
    \textbf{Extracted illumination from relit images.} We use our relighting model to relight a spherical light probe with varying illumination, and recover the environment maps from the relit images. Notice that while the content of the environemnt map is preserved, we can observe some warping inconsistencies between samples. This motivates our solution of using per-image shading embeddings to accommodate for this type of inconsistencies.
    }
    \label{fig:envmap}
\end{figure}
To reconstruct a highly reflective object perfectly, we would need the reflections to be exactly consistent across all the input views. Otherwise, any inconsistencies would appear as flickering in the 3D reconstruction. However, we observe that while our relighting model preserves the reflection content, some of the inconsistencies can appear as warping in the reflected environment. In Figure~\ref{fig:envmap}, we relight a spherical light probe that we rendered using random poses with different environment maps. Each of the relit images can be treated as a separate light probe from which we can extract the illumination, and compare the different extracted environment maps in order to visualize the inconsistencies. Here we find that while the content is largely preserved, some objects like the house and the trees suffer from some distortion across the different relit views. This motivates our novel shading embeddings: it allows the model to optimize the radiance field under a single constant illumination condition, while optimizing for per-image surface normals used for rendering reflections.

\section{Additional 3D Reconstruction Details}

This section provides additional details on our 3D reconstruction approach described in Section~3.2 of the main paper. Our method is based on NeRF-Casting~\cite{verbin2024nerf} with a few modifications.

First, NeRF-Casting is designed for real scenes without masks. In order to apply it to our setting where background content is masked out with white pixels, we replace its ``reflection features'', which volume renders a field of features along the reflected ray $\mathbf{o}'+t\mathbf{d}'$, with a single feature queried infinitely far away, \ie, at: 
\begin{equation}
\lim_{t\rightarrow\infty} \mathcal{C}\left(\mathbf{o}'+t\mathbf{d}'\right)=2\mathbf{d}',
\end{equation}
where $\mathcal{C}$ is the contraction function from Zip-NeRF~\cite{barron2023zipnerf}:
\begin{align}
    \mathcal{C}(\mathbf{x})=\begin{cases}\mathbf{x} \quad &\text{if } \|\mathbf{x}\| \leq 1, \\\left(2 - \frac{1}{\|\mathbf{x}\|}\right)\frac{\mathbf{x}}{\|\mathbf{x}\|} \quad &\text{if } \|\mathbf{x}\| > 1. \end{cases}
\end{align}

This simplifies the derivation in Section 4.2 of NeRF-Casting, so that the feature corresponding to the reflected ray $\mathbf{p}'$ is:
\begin{equation}
    \bar{\mathbf{f}} = \frac{1}{K}\sum_{j=1}^K \mathbf{f}(2\mathbf{d}'_j)\odot \operatorname{erf}\left((\sqrt{8}\boldsymbol{\nu}\sigma)^{-1}\right),
\end{equation}
where $\{\mathbf{d}'_j\}_{j=1}^K$ are NeRF-Casting's $K$ unscented reflection directions, $\sigma=2\gamma(\dot{r}+\bar{\rho})$ is the  scaling parameter defined in NeRF-Casting for infinitely distant content, $\boldsymbol{\nu}$ is a vector (with the same dimension as $\mathbf{f}$) containing the scale of NGP grid resolutions, $\odot$ denotes elementwise multiplication. See NeRF-Casting for additional information.

We also make a few additional small modifications to NeRF-Casting's optimization:
\begin{enumerate}
\item We optimize our NeRF for $25$K iterations rather than $50$K. We use the same learning rate schedule as in NeRF-Casting.
\item We initialize density around $\exp(-1)$ instead of $\exp(2)$.
\item We use a faster coarse-to-fine rate: using the notation from Appendix C.1. in~\cite{verbin2024nerf}, we set $m=16$ and $s=50$.
\item We remove the view direction as input into the color prediction network.
\end{enumerate}

Finally, for scenes from NAVI~\cite{jampani2023navi}, which have impercise camera poses, we found that adding a simple mask loss improved our results. For a ray with rendering weights $\{w_i\}_{i=1}^N$ we use the following loss:
\begin{equation} \label{eq:maskloss}
    \mathcal{L}_{\text{mask}} = \lambda_{\text{mask}}\cdot \left(\sum_{i=1}^N w_i - \alpha\right),
\end{equation}
where $\sum_{i=1}^N w_i$ is the opacity of the ray, and $\alpha$ is $1$ for object pixels and $0$ for background ones. Since the object masks provided with NAVI are also imprecise, we do not apply the mask loss in Equation~\ref{eq:maskloss} to pixels that are within $7$ pixels from a boundary. In our NAVI experiments we set $\lambda_{\text{mask}}=0.01$.

\section{Ablation Study Details}
\topic{Number of sampled frames}
To compare the effects of the number of frames the model relights simultaneously, we fine-tune our final model to relight 1 frame, 8 frames, 16 frames, and 32 frames at once. Then, using each model, we sequentially relight the entire 64-frame input under the same reference image. Our hypothesis is that with fewer frames, the relit outputs would be less consistent, and exhibit a drop of performance in the reconstruction. This was indeed the case, as showed in Table~3 of the main paper. We find that we gain a significant boost in performance going from single-frame relighting (what prior work follows) to 8-frame relighting, and additional gain as we increase the number of relit frames to 64. However, as expected, we notice diminishing returns where the benefit become more subtle as we increase the number of frames we relight at once.

\topic{Dataset ablation}
To investigate the benefit of augmenting our training data with highly reflective materials, we trained two models: one model only on standard assets, and one model where we randomly sample from standard assets, and assets augmented with highly reflective materials. For the sake of efficiency, we only train a 16-frames version of the model, and we train each model for $70$K training steps. We find that adding highly reflective assets significantly improves the model's performance on shiny assets, and surprisingly that it also provide a benefit to standard assets. Intuitively, this can be attributed to the fact that shiny assets are significantly more challenging to relight, and are more beneficial to the improvement of the model's performance than diffuse objects.

\topic{Shading embedding ablations}
While in Section~\ref{sec:shading} we motivate shading embeddings visually, we demonstrated its importance by comparing our novel shading embeddings and the standard appearance embeddings that prior work used. As we also show in Section~5 of the main paper and in the supplementary webpage, we show that the typical appearance embeddings are worse than not using any appearance embeddings and training on the relit images directly. This can be attributed to the fact that appearance embeddings can absorb any view-dependent changes that are necessary for realistic reflections, and render a mostly-diffuse object. On the other hand, not using any appearance embeddings can allow the reflections to move naturally along the object, but also include the flickering from the inconsistencies. Our shading embeddings resolve these issues: they can explain away any inconsistencies due to the reflections warping, while preserving the view-dependent effects necessary to render realistic reflections.


\end{document}